\documentclass[sigconf]{acmart}

\usepackage{etoolbox}

\usepackage{graphicx}

\usepackage{latexsym}
\usepackage[T1]{fontenc}
\usepackage[utf8]{inputenc}
\usepackage{microtype}

\usepackage{graphicx}
\usepackage{booktabs}
\usepackage{multirow}
\usepackage[most]{tcolorbox}
\usepackage{listings}
\usepackage{xcolor}
\usepackage{tcolorbox}
\usepackage{float}
\usepackage{enumitem}
\usepackage{stfloats}
\usepackage{placeins}
\usepackage[most]{tcolorbox}
\usepackage{subcaption}
\usepackage[ruled,vlined]{algorithm2e}
\usepackage{booktabs}
\usepackage{graphicx}
\usepackage{colortbl}
\usepackage{multirow}
\usepackage{array}

\definecolor{BudgetBlue}{RGB}{0,114,178}
\definecolor{ThoughtColor}{RGB}{0,92,175}
\definecolor{ActionColor}{RGB}{213,94,0}
\definecolor{ObservationColor}{RGB}{0,128,92}
\usepackage[ruled,vlined]{algorithm2e}

\newcommand{\Thought}{\mbox{\textcolor{ThoughtColor}{\emph{Thought}}}}
\newcommand{\Action}{\mbox{\textcolor{ActionColor}{\emph{Action}}}}
\newcommand{\Observation}{\mbox{\textcolor{ObservationColor}{\emph{Observation}}}}
\newcommand{\PhaseLine}[1]{%
  \textcolor{BudgetBlue!65!black}{\ttfamily // #1}\;
}

\AtBeginDocument{%
  }

\begin{document}

\settopmatter{printacmref=false}
\setcopyright{none}
\renewcommand\footnotetextcopyrightpermission[1]{}
\pagestyle{plain}
\title{StepKV: Step-Aware KV Cache Compression for LLM Agents}

\author{%
  Boyu Feng$^{1}$, 
  Jiahong Liu$^{1}$, 
  Yifan Li$^{1}$,
  Wenhao Yu$^{1}$,
  Zexuan Qiu$^{1}$,
  Yuliang Sun$^{1}$,
  Ming Shen$^{1}$,
  Xiang Li$^{2}$,
  Quanyu Dai$^{2}$,
  Irwin King$^{1}$\\[0.5em]
  $^{1}$The Chinese University of Hong Kong\\
  $^{2}$Huawei Technologies Co., Ltd\\[0.3em]
}

\begin{abstract}
Key-value (KV) caching is essential for efficient autoregressive large language model (LLM) inference, but the cache grows linearly with the accumulated context and increases the amount of KV data that must be stored and read during decoding. \textbf{KV cache compression} mitigates this cost by retaining only a subset of cached tokens. This problem becomes especially important in multi-step LLM agents, where a user query expands into a trajectory of intermediate reasoning, tool interactions, and retrieved observations. Existing pruning methods usually treat the cache as a flat token stream and rank tokens by recency or attention-based saliency. This creates a mismatch between the unit of compression and the unit of reasoning: token-level pruning removes individual cache entries, whereas useful information in multi-step agents is often organized by reasoning steps whose importance is uneven and delayed. As a result, an early observation or intermediate decision may receive little recent attention but become necessary for later evidence synthesis. This failure mode is termed \textbf{\emph{Reasoning Continuity Disruption}}.
These observations motivate KV cache compression that operates jointly at the token and reasoning-step levels.\textbf{StepKV} addresses this goal by treating reasoning steps as first-class retention units. Instead of allocating cache budget only among isolated tokens, the framework associates cache entries with their generating steps, estimates step utility from \textbf{trajectory-derived signals}, and combines this utility with token-level saliency. The resulting score globally ranks all prunable tokens, from which StepKV retains the top-scoring entries under the target budget. More broadly, StepKV provides a \textbf{step-centric perspective} for agent KV cache compression.Across agent multi-hop QA and long-horizon web reasoning tasks, StepKV sustains accuracy under low KV budgets where token-level baselines degrade sharply, offering a more robust \textbf{efficiency–accuracy trade-off} for multi-step agent inference.
\end{abstract}

\maketitle
\begingroup
\renewcommand{\thefootnote}{}
\footnotetext{E-mail: \texttt{\{fengboyu, jhliu\}@link.cuhk.edu.hk}.
Corresponding author: Jiahong Liu. }
\endgroup

\begin{figure}[!t]
    \vspace{0.5cm}
    \hspace{-0.5cm}
\includegraphics[width=1.05\linewidth, trim={12mm 6mm 9mm 3mm}, clip]{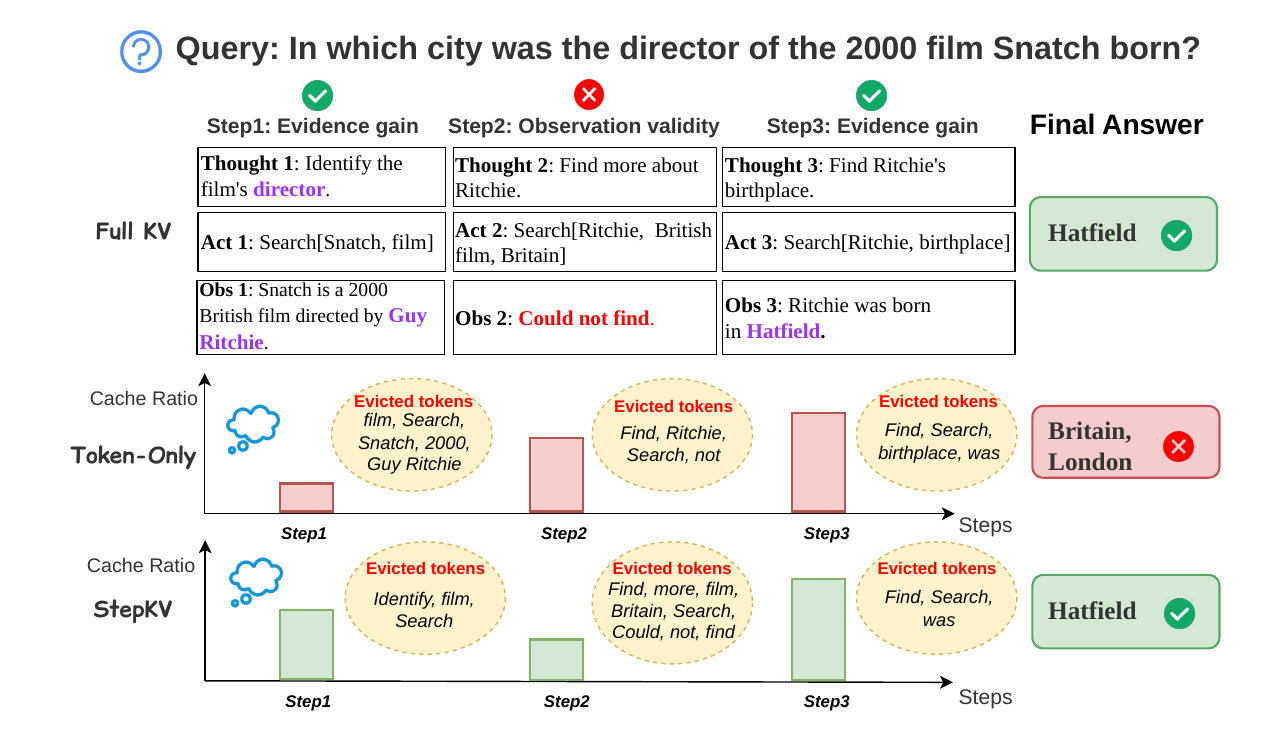}
\caption{Comparison of StepKV and token-level KV cache pruning in preserving critical reasoning steps.}
    \label{fig:case}
    \vspace{-0.55cm}
\end{figure}

\section{Introduction}
Key-value (KV) caching is a standard mechanism for efficient autoregressive large language model (LLM) inference~\cite{vaswani2017attention,pope2022efficiently}. By storing the keys and values of previous tokens, the model avoids recomputing past hidden states when generating each new token. This acceleration, however, comes with a growing memory and data-movement cost: the cache expands with the accumulated context, and decoding must repeatedly read historical KV entries. \textbf{KV cache compression} therefore becomes a practical tool for reducing memory footprint and improving the efficiency of long-context inference~\cite{zhang2023h2o,li2024snapkv,xiao2024streamingllm}.

This need is amplified in \textbf{LLM agents}~\cite{yao2023react,schick2023toolformer,nakano2022webgpt,shinn2023reflexion}. Unlike single-pass generation, agentic inference often expands one user query into a trajectory of intermediate reasoning, tool interactions, and retrieved observations. In multi-hop question answering and web-based information seeking, these trajectories can contain evidence collected at different stages of the process. Compression must therefore reduce the cache while preserving information that later reasoning steps may still need.

Existing KV cache pruning methods usually treat the cache as a \textbf{flat token stream}~\cite{zhang2023h2o,ge2024modeltells,li2024snapkv,wu2025scope}. They score individual tokens by recency, accumulated attention, or related saliency signals, then keep the tokens that appear locally important. This token-level view is effective in many long-context settings, but it does not explicitly model the structure that produced the cache. In step-structured agents, the \textbf{unit of compression can diverge from the unit of reasoning}: useful information is often introduced and reused at the level of a reasoning step, not only at the level of isolated tokens.

The resulting failure mode is \textbf{\emph{Reasoning Continuity Disruption}}. Under tight KV budgets, token-level pruning may remove early observations or intermediate decisions before their later reuse. The agent can then lose evidence needed to connect later observations, causing repeated tool calls, missing evidence, or incorrect final answers. This points to KV cache compression for multi-step agents that preserves trajectory structure rather than only locally salient tokens.

To address the mismatch between existing KV cache compression methods and the structured nature of agent reasoning, \textbf{StepKV} introduces reasoning-step importance into token-level cache selection through a \textbf{step-aware token scoring mechanism}. 
Unlike conventional autoregressive generation, agent reasoning is typically composed of multiple consecutive reasoning steps, where different steps may serve distinct functions, such as information acquisition, hypothesis formation, intermediate derivation, and final decision making. 
Therefore, different reasoning steps contribute unequally to future inference and final outcomes, and estimating importance solely from individual tokens may overlook intermediate steps with long-term impacts on the reasoning trajectory.

Figure~\ref{fig:case} presents an example to illustrate the mismatch between token-level importance estimation and reasoning-step-level importance in agent inference. 
The upper part of Figure~\ref{fig:case} illustrates an important phenomenon observed during agent reasoning: different reasoning steps contribute unequally to the final outcome. 
Some steps provide critical intermediate information that directly affects subsequent reasoning, while others mainly contain redundant exploration or auxiliary content. 
However, existing token-level importance metrics, such as attention-based scores, may not fully capture this semantic difference between reasoning steps. 
As shown in the upper part of Figure~\ref{fig:case}, tokens from less critical steps may still receive considerable attention or importance scores due to their local relevance, causing token-level compression methods to preserve locally salient but globally less useful information. 
Meanwhile, tokens that are essential for maintaining future reasoning dependencies may be mistakenly discarded.

The lower part of Figure~\ref{fig:case} demonstrates how StepKV alleviates this issue. 
By incorporating step-level importance into token selection, StepKV can distinguish the different contributions of reasoning steps and adjust cache retention accordingly. 
Tokens belonging to high-utility reasoning steps receive higher priority during cache selection, while redundant information from less important steps can be removed more effectively. 
This case demonstrates that reasoning-step awareness provides complementary information beyond token-level scoring and enables more effective KV cache management for long-horizon agent reasoning.

Based on this observation, \textbf{StepKV} does not treat reasoning steps as coarse-grained retention units. 
Instead, it utilizes step-level utility to guide token-level importance estimation. 
The final cache selection jointly considers local token saliency and the global contribution of the corresponding reasoning step, allowing critical tokens from important steps to receive higher retention priority while maintaining the flexibility of token-level compression. 
This step-aware perspective better aligns KV cache management with the trajectory structure of agent reasoning and effectively mitigates \textbf{Reasoning Continuity Disruption} in long-horizon inference.
Overall, our contributions can be summarized as follows:

\begin{itemize}

    \item \textbf{A trajectory-aware view of KV cache compression for LLM agents.} The paper reframes KV cache compression as a problem of preserving useful reasoning trajectories under fixed cache budgets.
    
    \item \textbf{A reasoning-continuity-preserving compression framework.} StepKV provides a step-aware cache allocation strategy that protects cross-step dependencies while retaining token-level selectivity.
    
    \item \textbf{Empirical support for trajectory-aware KV compression.} Experiments show that trajectory-aware compression provides a stronger \textbf{efficiency--accuracy trade-off} than token-level pruning under tight KV budgets.

\end{itemize}

\section{Related Work}

\subsection{LLM Agents}
LLM agents extend single-pass generation into multi-step trajectories that interleave reasoning, tool use, and environmental feedback~\cite{yao2023react,schick2023toolformer,nakano2022webgpt,shinn2023reflexion}. Unlike conventional generation, agent inference continuously accumulates thoughts, actions, observations, and retrieved evidence, whose utility may emerge only in later steps. Most existing studies focus on planning, tool use, retrieval, and memory design, while recent work has begun to manage the inference context or KV cache according to model directives, trajectory structure, or user intent~\cite{kariyappa2026sidequest,ma2026leyline,li2026intentkv}. StepKV complements these efforts by addressing fine-grained retention within a fixed KV cache budget. It explicitly models the semantic boundaries and utility of reasoning steps while retaining salient tokens within each step, thereby preserving reasoning continuity under aggressive cache compression.

\subsection{KV Cache Management}
Long-context inference has been optimized through architectural changes, sparse attention, and explicit KV cache compression. GQA and MLA reduce the memory footprint of KV representations~\cite{ainslie2023gqa,deepseekai2024deepseekv2}, while sparse-attention and state-space architectures reduce the computation required for long-range dependency modeling~\cite{yuan2025nativesparse,gu2024mamba,yang2025gateddelta}. These approaches primarily optimize how contextual representations are computed or stored. Decode-stage cache management instead determines which historical KV entries to retain as generation proceeds. Representative methods estimate token importance from accumulated attention statistics~\cite{zhang2023h2o,scissorhands},  query-aware attention~\cite{tova,li2024snapkv}, or structural signals~\cite{tokenskipping,xiao2024streamingllm}. Recent methods further incorporate redundancy- or reasoning-aware signals into cache reduction~\cite{cai2026rkv,hu2025raas,zhang2025lazyeviction,ramachandran2026thinkv,kariyappa2026sidequest}. 
However, these methods still primarily operate at the token level and overlook the structured nature of agent reasoning trajectories. 
StepKV introduces step-aware cache allocation for agent reasoning, preserving critical agent trajectories while removing redundant context. 
For detailed discussion of related works, please refer to Appendix~\ref{app:relatedworks}.

\section{Problem Formulation}
\label{sec:problem}

\noindent Given a user query~$q$ and an external environment, let $\mathrm{Step}_k$ denote the $k$-th completed interaction step of an LLM agent. After step~$t$ is completed, the finalized steps are indexed by $\mathcal{S}^{(t)}=\{1,\ldots,t\}$ and form the text trajectory $\tau^{(t)}=(q,\{\mathrm{Step}_k\}_{k\in\mathcal{S}^{(t)}})$. At step~$k$, the model generates the \Thought{} and \Action{} fields, executes the \Action{} in the environment, and appends the returned \Observation{} to the context. We write $\mathrm{Th}_k$, $\mathrm{Act}_k$, and $\mathrm{Obs}_k$ for these three fields, respectively.

\textbf{Cache compression} operates on the tokenized trajectory after step~$t$ has been appended. Let $\mathcal{I}^{(t)}=\{1,\ldots,N_t\}$ index the prunable trajectory tokens, excluding protected prompt/query tokens. Each completed step occupies a contiguous span in $\mathcal{I}^{(t)}$. Let $\mathcal{C}^{(t)}$ denote the KV cache at this post-step pruning boundary. It consists of protected entries $\mathcal{C}_{\mathrm{prot}}$ and prunable trajectory entries $\mathcal{C}_{\mathrm{traj}}^{(t)}$, where $\mathcal{C}_{\mathrm{traj}}^{(t)}[i]$ is the KV entry for token~$i\in\mathcal{I}^{(t)}$. Thus, $\tau^{(t)}$ is the text trajectory, whereas $\mathcal{I}^{(t)}$ is the token-index set over that trajectory. Given a keep ratio~$\rho$, online cache compression selects a retained index set $\mathcal{K}^{(t)}\subseteq\mathcal{I}^{(t)}$ under a fixed budget. We use $\|$ to denote cache concatenation, and $\mathcal{C}_{\mathrm{traj}}^{(t)}[\mathcal{K}^{(t)}]$ to denote the trajectory-cache entries selected by $\mathcal{K}^{(t)}$:
\begin{equation}
\label{eq:problem}
\begin{aligned}
\mathcal{C}^{(t)}
&=\mathcal{C}_{\mathrm{prot}}
\;\|\;
\mathcal{C}_{\mathrm{traj}}^{(t)}
\quad\text{\footnotesize\textcolor{black!65}{(full KV)}},\\
\mathcal{K}^{(t)}
&\subseteq\mathcal{I}^{(t)},
\qquad
|\mathcal{K}^{(t)}|\le
B_t=\max(1,\lfloor \rho N_t \rfloor),\\
\widehat{\mathcal{C}}^{(t)}
&=
\mathcal{C}_{\mathrm{prot}}
\;\|\;
\mathcal{C}_{\mathrm{traj}}^{(t)}[\mathcal{K}^{(t)}]
\quad\text{\footnotesize\textcolor{black!65}{(compressed KV)}}.
\end{aligned}
\end{equation}
\textbf{The compression objective is to choose $\mathcal{K}^{(t)}$ so that $\widehat{\mathcal{C}}^{(t)}$ satisfies the target cache budget while preserving the agent's task-level performance relative to full-cache inference with $\mathcal{C}^{(t)}$.} In other words, under a fixed keep ratio, the compressed agent aims to maintain the final answer quality or task success of the original full-cache run as closely as possible.

\begin{figure*}[t]
    \centering

    \begin{subfigure}[c]{0.31\textwidth}
        \centering
        \includegraphics[
            width=\linewidth,
            height=5cm,
            keepaspectratio
        ]{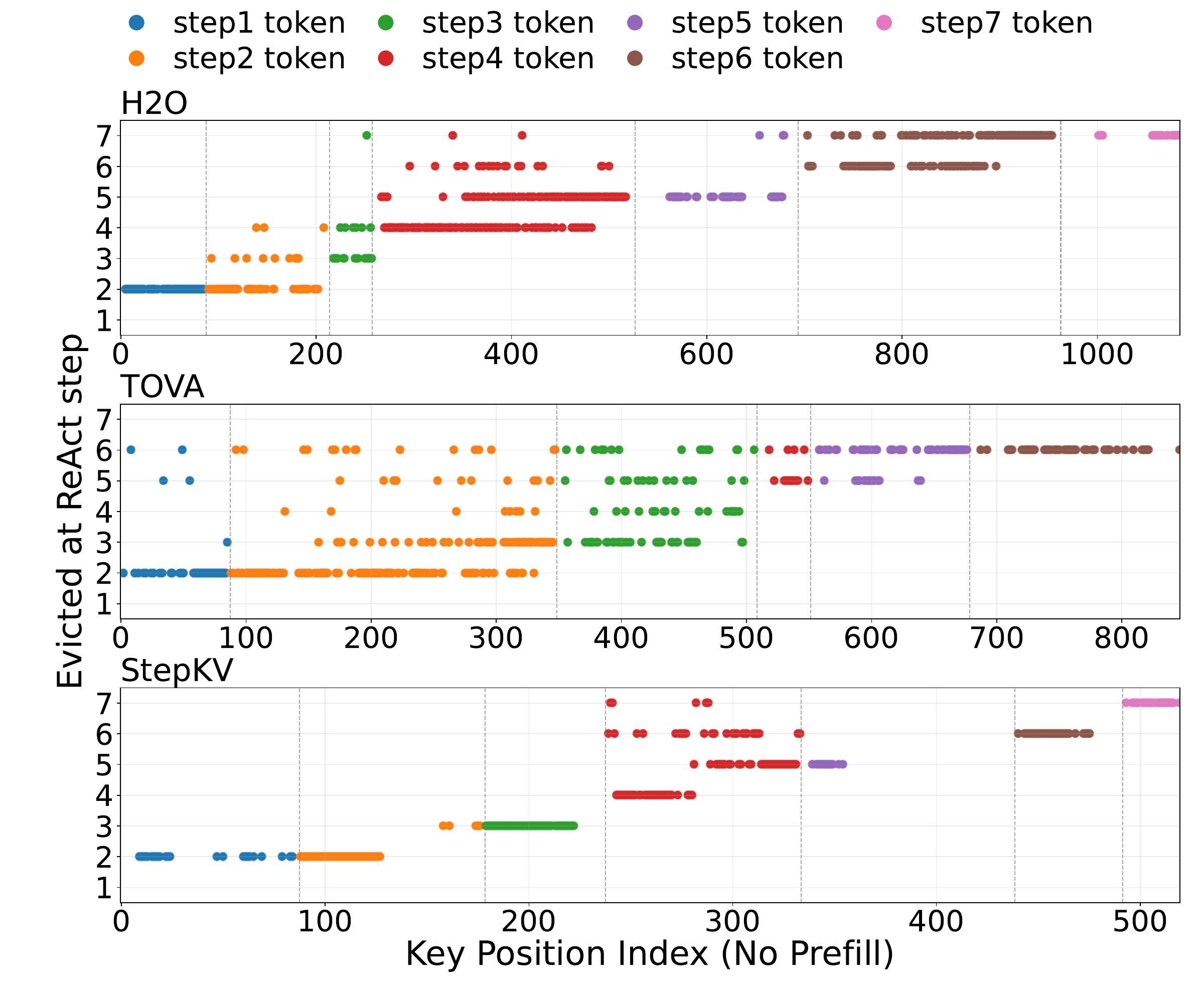}
        \caption{Evicted tokens by originating step.}
        \label{fig:step_drop}
    \end{subfigure}\hspace{0.015\textwidth}%
    \begin{subfigure}[c]{0.34\textwidth}
        \centering
        \includegraphics[
            width=\linewidth,
            height=5cm,
            keepaspectratio
        ]{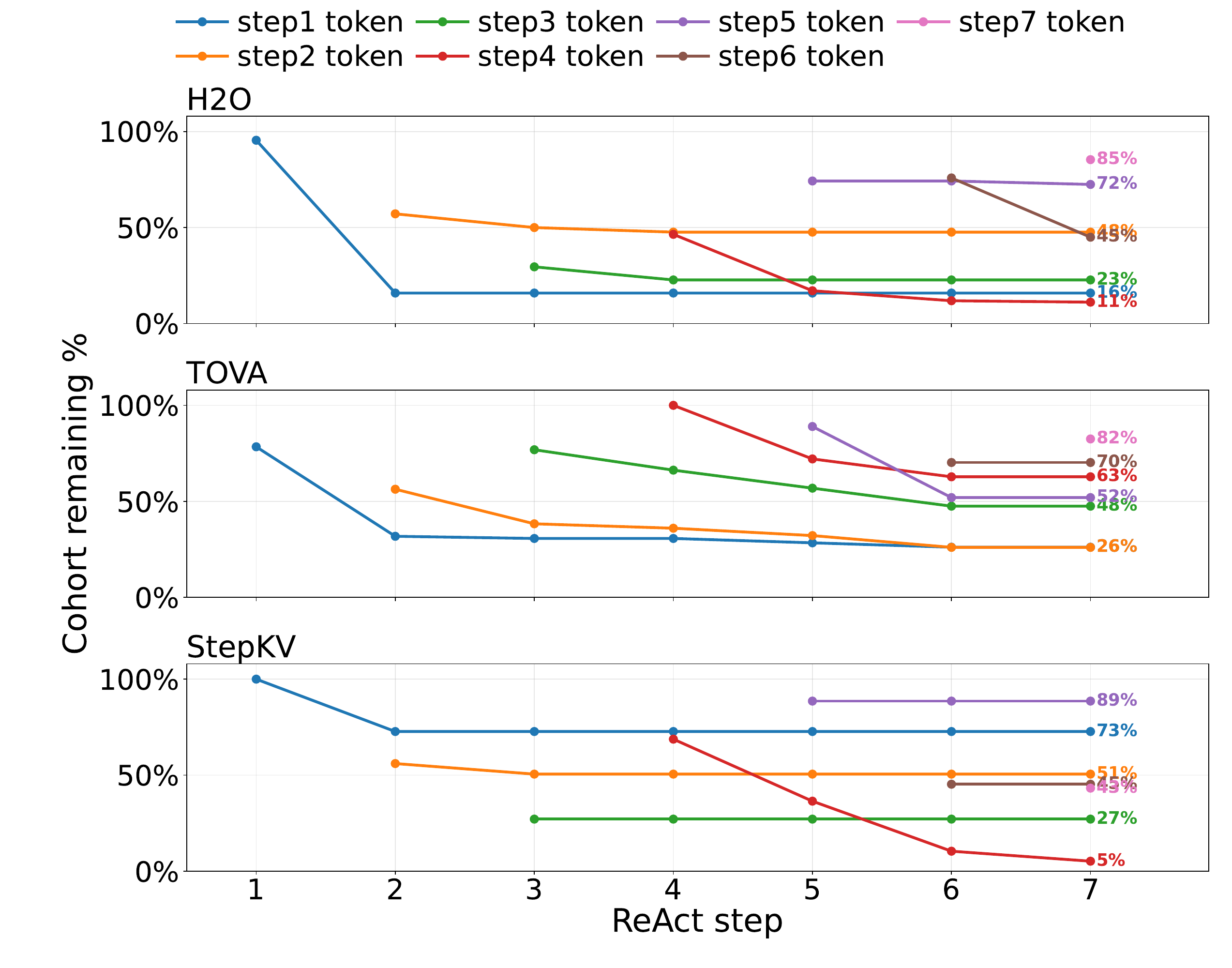}
        \caption{Remaining step cohorts.}
        \label{fig:cohort_step_token}
    \end{subfigure}\hspace{0.015\textwidth}%
    \begin{subfigure}[c]{0.32\textwidth}
        \centering
        \includegraphics[
            width=\linewidth,
            height=5cm,
            keepaspectratio
        ]{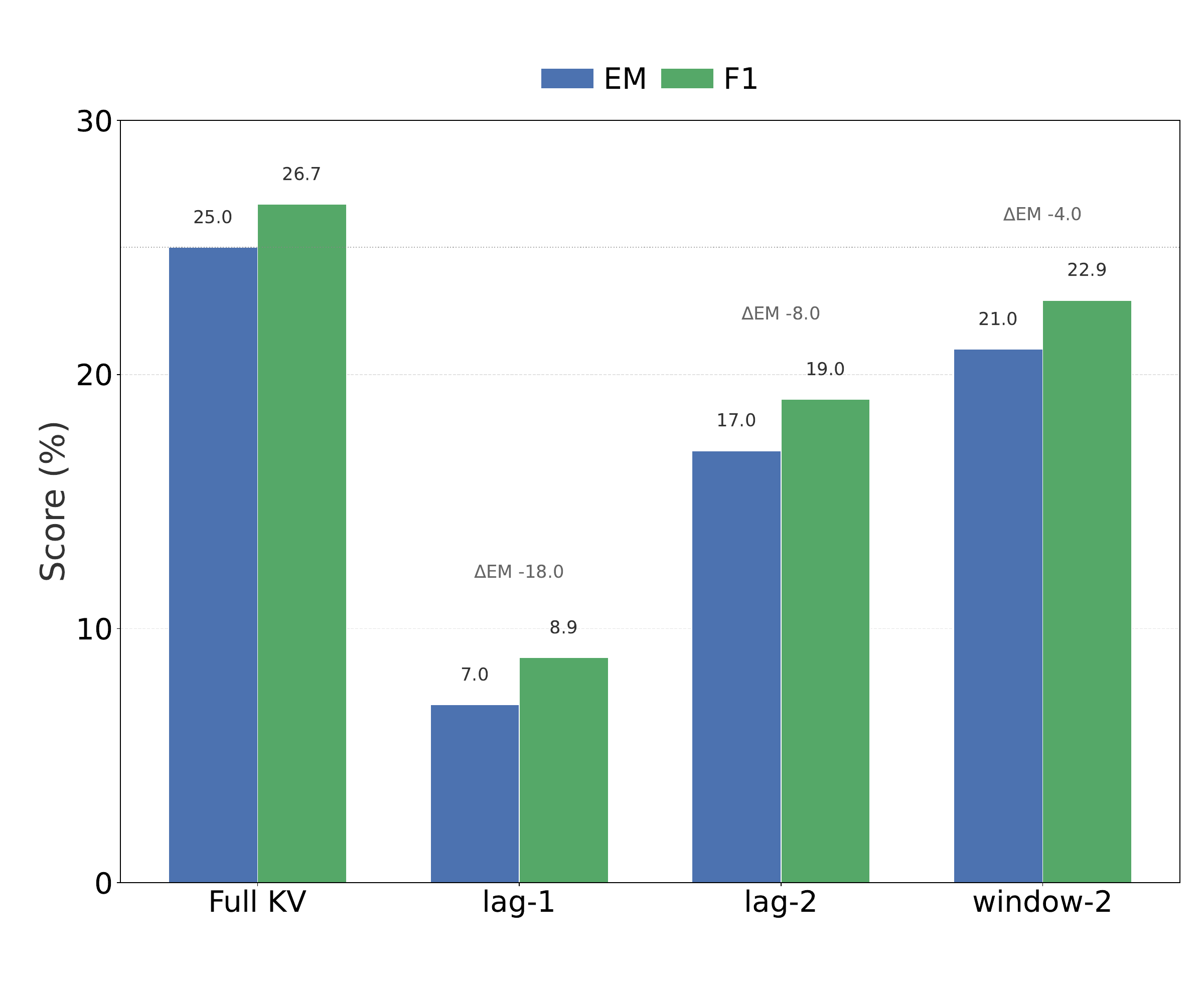}
        \caption{Performance under step removal.}
        \label{fig:step_removal}
    \end{subfigure}

    \caption{Motivation for reasoning continuity under KV cache pruning. Panels (a)--(b) trace a representative rollout after prompt prefill: (a) groups evicted tokens by their source step, showing that pruning spans the preceding trajectory, and (b) tracks each cohort's remaining fraction, showing that even recently completed steps can be rapidly depleted. Panel (c) reports EM and F1 under controlled history removal. \texttt{lag-1} and \texttt{lag-2} remove the completed step one or two positions before the current step, respectively, while \texttt{window-2} retains the two most recent completed steps and removes all earlier ones. Every removal setting degrades performance, but the magnitude depends strongly on which part of the trajectory is removed.}
    \label{fig:step_analysis}

    \vspace{-0.4cm}
\end{figure*}

\section{Motivating Observation: Reasoning Continuity Disruption}
\label{sec:motivation}
The problem formulation exposes a structural tension between the unit of pruning and the unit of reasoning. KV cache compression selects individual cache entries, but an agent trajectory is produced through completed reasoning steps, each containing intermediate decisions, tool calls, and observations. Therefore, the relevant question is not only whether high-saliency tokens are retained, but whether the retained cache still contains a usable trace of the steps that later reasoning may need. Figure~\ref{fig:step_analysis} highlights two observations behind this mismatch.

\textbf{Motivation experiment.}
To make these observations visible, Figure~\ref{fig:step_analysis} combines a trace diagnostic with a controlled step-removal study. For Figures~\ref{fig:step_analysis}(\subref{fig:step_drop}) and~(\subref{fig:cohort_step_token}), a representative agent rollout is instrumented at the token and step levels. Each generated token after prompt prefill is assigned to its source step; H$_2$O and TOVA are then applied online under the same target cache budget, and every pruning decision records which step each retained or evicted token came from. Figure~\ref{fig:step_analysis}(\subref{fig:step_removal}) reruns agent inference under otherwise identical settings while removing the step at lag one or lag two, or deleting all history older than the two most recent completed steps. The trace diagnostic first identifies which step context token-level methods actually remove; the controlled removal then tests whether losing such context directly changes downstream EM and F1.

\noindent\textbf{Observation 1: Token-level pruning discards both early and recent step context.}
H$_2$O and TOVA make retention decisions from attention-derived importance at the token level; neither method explicitly tracks whether a completed reasoning step remains sufficiently represented after eviction. We analyze how these methods evict tokens across completed steps as an agent trajectory unfolds. \textit{First, early-step information is easily discarded.} Figure~\ref{fig:step_analysis}(\subref{fig:step_drop}) shows that evictions repeatedly span earlier steps as the trajectory advances, leaving only fragmented traces of their reasoning, actions, and observations. \textit{Second, recently completed steps can be depleted almost immediately.} In Figure~\ref{fig:step_analysis}(\subref{fig:cohort_step_token}), a new step cohort can lose a large fraction of its tokens after only one or two subsequent steps. Token-level pruning can therefore remove both long-range evidence from early steps and the nearby context required for the next reasoning transition, even while satisfying the global cache budget.

\noindent\textbf{Observation 2: Loss of step context directly degrades downstream reasoning.}
Figure~\ref{fig:step_analysis}(\subref{fig:step_removal}) verifies that the deletion behavior in Figures~\ref{fig:step_analysis}(\subref{fig:step_drop}) and~(\subref{fig:cohort_step_token}) is not a benign reduction of redundant tokens. Every controlled removal setting underperforms Full KV. \texttt{window-2} removes all but the two most recent completed steps and still reduces both EM and F1, showing that earlier steps continue to supply useful evidence. Removing the immediately preceding step with \texttt{lag-1} causes the largest degradation, while removing the step two positions back with \texttt{lag-2} is also substantially worse than retaining both recent steps. The step context fragmented by token-level eviction therefore directly affects downstream task quality. Moreover, the unequal drops show that the required context changes with the trajectory: a new step can be critical to the next transition, while an older step remains important when its evidence is reused later.

These observations identify \textbf{\emph{Reasoning Continuity Disruption}}: token-only importance can satisfy the cache budget while breaking the step structure required by downstream reasoning. The missing signal is a \textbf{step score} that assigns shared retention importance to tokens from the same completed step. Because the usefulness of a step changes as later dependencies emerge, this score also needs to be initialized from the completed step and updated as subsequent trajectory content reuses its information. Section~\ref{sec:methodology} instantiates this step-level signal while preserving token-level selectivity.

\vspace{-0.5cm}
\begin{figure*}[h]
    \centering
    \includegraphics[width=\linewidth]{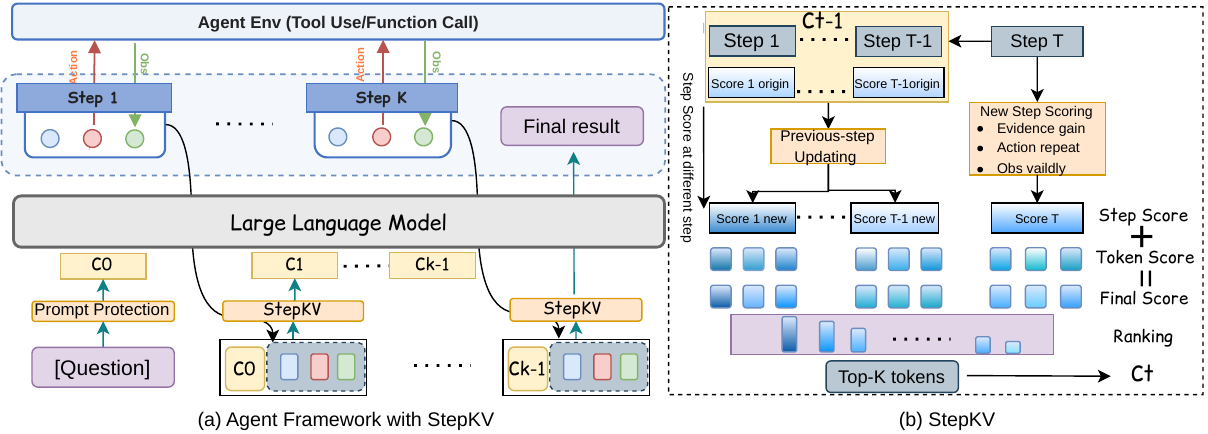}
    \caption{Overview of StepKV. (a) The integration of StepKV into the LLM agent inference framework. (b) The StepKV pipeline.}
    \label{fig:method}
    \vspace{-0.2cm}
\end{figure*}
\vspace{0.2cm}
\section{Methodology: Step-Aware KV Cache Compression}
\label{sec:methodology}

With the cache layout and step spans defined in Section~\ref{sec:problem},
StepKV addresses the granularity mismatch between KV cache compression and
step-structured agent reasoning. Instead of making retention decisions solely
at the token level, StepKV introduces a trajectory-level utility signal $S_k$
for each completed step and updates it when later reasoning reveals additional
dependencies. Meanwhile, token saliency $T_i$ preserves fine-grained selection
within each step. These two signals jointly determine the retention priority
$P_i$ of each token, and the top $B_t$ tokens form the retained set
$\mathcal{K}^{(t)}$.

The key design is to allocate cache based on completed reasoning steps while
maintaining token-level selectivity. When a new step completes, its immediate
contribution initializes the step utility (\textit{Phase 1}). As the
trajectory evolves, later reuse updates previous step utilities through
delayed dependency signals (\textit{Phase 2}). Finally, the step utility is
combined with token saliency for cache selection (\textit{Phase 3}). This
design enables cache compression to follow the evolving dependency structure
of agent reasoning rather than relying only on instantaneous token-level
signals.

\textbf{Design Rationale.}
The step utility is motivated by the temporal dependency of agent reasoning.
The value of an intermediate step is not always observable when it is
generated: some steps provide immediate useful evidence, while others become
important only when later steps reuse their information. Therefore, StepKV
decomposes step utility into an immediate contribution term $r_k$ and a
delayed reuse accumulator $c_k$.In addition, they provide a trajectory-level
retention signal that complements token saliency and captures both newly
introduced information and future dependencies. Additional analysis of these
signals is provided in Appendix~\ref{step_score_analysis}.

\subsection{Step Utility Scoring}
\label{sec:step-utility}

\textbf{A useful reasoning step contributes meaningful evidence to the trajectory, and its importance may increase when later reasoning depends on that evidence.} This follows directly from Section~\ref{sec:motivation}: reasoning steps contribute unequally, while the value of early information may become apparent only after subsequent tool calls. Step importance therefore cannot be determined once and then kept fixed. StepKV models this evolving importance through two complementary operations. \emph{New-step scoring} estimates the immediate progress made by a newly completed step, whereas \emph{previous-step updating} revises an earlier step when its evidence is reused. These operations produce a completion score $r_k$ and a reuse accumulator $c_k\ge0$, which are combined into the step utility $S_k$.

\vspace{-0.2cm}
\subsubsection{New-step Scoring ($r_k$)}
When step~$k$ is finalized, only its immediate contribution is observable. Rather than treating the completed step as undifferentiated text, StepKV distinguishes the roles of the \Thought{}, \Action{}, and \Observation{} fields as reasoning intent, environment interaction, and returned evidence, respectively. These roles yield three field-level signals.

\noindent(1) \textbf{Observation validity.}
The validity of the returned observation is defined as:
\begin{equation*}
\mathrm{valid}_k=
\begin{cases}
1, & \text{if $\mathrm{Obs}_k$ is usable},\\
0, & \text{otherwise}.
\end{cases}
\end{equation*}
\textit{Remark:} An empty, invalid, or error-returning $\mathrm{Obs}_k$ is considered unusable and therefore contributes no positive evidence to the new-step score.

\noindent(2) \textbf{Evidence gain.}
Let $\operatorname{sim}(a,b)$ denote overlap-based textual similarity. Evidence gain compares the concatenated Thought--Observation content of the current step with that of each previous step:
\begin{equation*}
\mathrm{gain}_k
=
1-
\max_{1\le u<k}
\operatorname{sim}
\!\left(
\mathrm{Th}_k\Vert\mathrm{Obs}_k,
\mathrm{Th}_u\Vert\mathrm{Obs}_u
\right),
\qquad \mathrm{gain}_1=1.
\end{equation*}
\textit{Remark:} The maximum identifies the closest earlier step, so the score is high when the current reasoning and evidence differ from what has already appeared. Observation validity and evidence gain are computed independently and enter $r_k$ as separate signals.

\noindent(3) \textbf{Action redundancy.}
Action redundancy is computed as:
\begin{equation*}
\mathrm{red}_k
=
\max_{1\le u<k}\operatorname{sim}(\mathrm{Act}_k,\mathrm{Act}_u),
\qquad \mathrm{red}_1=0.
\end{equation*}
\textit{Remark:} The score compares the current Action with all earlier actions and uses the strongest match as its redundancy penalty. Thus, repeated tool calls lower the contribution of a step unless they return useful evidence captured by the positive terms.

The three signals are combined into the initial score of step~$k$:
\begin{equation}
\label{eq:reward}
r_k
=
\mathrm{valid}_k
+
\mathrm{gain}_k
-
\lambda\,\mathrm{red}_k.
\end{equation}
The positive terms reward usable and relevant evidence, while $\lambda$ controls the penalty for repeated interaction.

\subsubsection{Previous-step Updating ($c_k$)}
The completion score cannot capture dependencies that emerge after a step is produced. StepKV therefore maintains a \textbf{reuse accumulator} $c_k$ for each completed step to record how strongly its evidence is reused later. \textbf{When step~$k$ is first completed, $c_k$ is initialized to zero; it is updated as subsequent steps reveal delayed relevance.} After step~$t$ is completed, StepKV compares the new observation $\mathrm{Obs}_t$ with each previous observation $\mathrm{Obs}_k$ and updates as:
\begin{equation}
\label{eq:reuse}
c_k
\leftarrow
c_k+
\min\!\left(
1,\;
\eta\,\operatorname{sim}(\mathrm{Obs}_t,\mathrm{Obs}_k)
\right),
\qquad 1\le k<t.
\end{equation}
\textit{Remark:} A previous step gains reuse credit only when its observation reappears later; zero overlap leaves $c_k$ unchanged, while $\eta$ and the cap control each increment.

\subsubsection{Step Utility ($S_k$)}
The utility of step~$k$ combines the completion score $r_k$ from new-step scoring with the reuse accumulator $c_k$ from previous-step updating:
\begin{equation}
\label{eq:step-score}
S_k=\operatorname{clip}\!\left(
w_r\,\operatorname{clip}(r_k,r_{\min},r_{\max})
+w_c\log(1+c_k),
\,0,\,
S_{\max}
\right),
\end{equation}
Here, $\operatorname{clip}(z,a,b)$ truncates $z$ to $[a,b]$. The completion term captures the step's immediate contribution, whereas $\log(1+c_k)$ adds delayed credit as later observations reuse its evidence. The weights $w_r$ and $w_c$ balance these two signals; the inner clip and logarithm control their scales, and the outer clip constrains $S_k$ to $[0,S_{\max}]$.

\subsection{Token Saliency and Cache Selection}
\label{sec:cache-selection}

StepKV complements step utility with a token saliency score $T_i$, which estimates the local importance of each prunable KV entry. Token saliency keeps the selector sensitive to model-level evidence, while $S_k$ preserves trajectory-level structure. This component is modular: $T_i$ can be supplied by existing token-level KV importance estimators or selection rules~\cite{zhang2023h2o,ge2024modeltells,li2024snapkv,wu2025scope} without changing the step-utility computation.
Each token inherits the utility of its source step. For token~$i$ produced by step~$k$, the retention priority is:
\begin{equation}
\label{eq:token-priority}
P_i=\alpha T_i+\beta S_k.
\end{equation}
Tokens in the same step share the same
$S_k$
but may differ in
$T_i$,
allowing locally salient tokens to survive even within a low-utility step, and high-utility steps to lift their entire span.

The final selector globally ranks tokens by $P_i$ and retains the top $B_t$. Step utility raises the priorities of tokens from important steps, while token saliency preserves within-step selectivity. No step is permanently protected: its tokens can still be evicted as $T_i$ is recomputed and $N_t$ grows under a fixed keep ratio.

\FloatBarrier

\vspace{-0.1cm}
\section{Experiments}
\label{sec:experiments}

We organize the evaluation around four research questions: \textbf{RQ1: Does StepKV preserve reasoning accuracy under tight KV budgets?} (Section~\ref{sec:rq1}); \textbf{RQ2: Does StepKV reduce practical cost in long-horizon agent reasoning?} (Section~\ref{sec:rq2}); \textbf{RQ3: How does reasoning depth affect compressed-cache performance?} (Section~\ref{sec:rq3}); and \textbf{RQ4: Do step-level signals explain the gains of StepKV?} (Section~\ref{sec:rq4}). 

\begin{table*}
\caption{Main results under different KV cache budgets. Full KV and ReAct are reference settings; bold numbers indicate the best compressed result for each model, budget, dataset, and metric.}
\centering
\footnotesize
\setlength{\tabcolsep}{4pt}
\renewcommand{\arraystretch}{1.08}
\begin{tabular*}{\textwidth}{@{\extracolsep{\fill}} l c cc cc cc}
\toprule
\multirow{2}{*}{Method} & \multirow{2}{*}{Budget} & \multicolumn{2}{c}{HotpotQA} & \multicolumn{2}{c}{2WikiMultihopQA} & \multicolumn{2}{c}{MuSiQue} \\
\cmidrule(lr){3-4} \cmidrule(lr){5-6} \cmidrule(lr){7-8}
& & EM~$\uparrow$ & F1~$\uparrow$ & EM~$\uparrow$ & F1~$\uparrow$ & EM~$\uparrow$ & F1~$\uparrow$ \\
\midrule
\rowcolor{gray!15}
\multicolumn{8}{c}{\textbf{Qwen2.5-7B-Instruct}} \\
Full KV
& 100\% & $26.40 \pm 1.73$ & $36.16 \pm 1.69$
& $24.20 \pm 8.47$ & $29.73 \pm 8.24$
& $5.13 \pm 0.76$ & $9.01 \pm 1.52$ \\
ReAct
& -- & $27.60 \pm 0.35$ & $38.25 \pm 0.81$
& $27.53 \pm 4.39$ & $33.77 \pm 4.46$
& $6.93 \pm 1.10$ & $13.98 \pm 1.07$ \\
\cmidrule{1-8}
H$_2$O
& \cellcolor{BudgetBlue!10}50\% & $12.00 \pm 0.72$ & $19.57 \pm 0.64$
& $10.67 \pm 1.42$ & $15.29 \pm 0.51$
& $1.93 \pm 0.42$ & $8.13 \pm 0.80$ \\
TOVA
& \cellcolor{BudgetBlue!10}50\% & $10.07 \pm 1.03$ & $19.05 \pm 2.06$
& $9.20 \pm 1.40$ & $15.87 \pm 1.58$
& $1.93 \pm 0.70$ & $8.07 \pm 0.28$ \\
TokenSkipping
& \cellcolor{BudgetBlue!10}50\%
& $8.80 \pm 2.28$
& $14.22 \pm 4.26$
& $6.87 \pm 0.41$
& $9.38 \pm 0.08$
& $1.13 \pm 0.17$
& $5.60 \pm 0.05$ \\
\textbf{StepKV (Ours)}
& \cellcolor{BudgetBlue!10}50\% & $\textbf{23.23} \pm 2.83$ & $\textbf{31.81} \pm 3.20$
& $\textbf{22.67} \pm 1.79$ & $\textbf{27.29} \pm 0.92$
& $\textbf{7.27} \pm 1.22$ & $\textbf{13.92} \pm 1.31$ \\
\cmidrule{1-8}
H$_2$O
& \cellcolor{BudgetBlue!20}20\% & $1.53 \pm 1.21$ & $3.52 \pm 1.01$
& $1.07 \pm 0.50$ & $2.64 \pm 0.73$
& $0.00 \pm 0.00$ & $0.94 \pm 0.41$ \\
TOVA
& \cellcolor{BudgetBlue!20}20\% & $0.73 \pm 0.58$ & $2.98 \pm 0.57$
& $1.47 \pm 0.99$ & $2.15 \pm 0.08$
& $0.07 \pm 0.12$ & $0.64 \pm 0.43$ \\
TokenSkipping
& \cellcolor{BudgetBlue!20}20\%
& $0.33 \pm 0.01$
& $0.39 \pm 0.03$
& $0.00 \pm 0.00$
& $0.08 \pm 0.01$
& $0.00 \pm 0.00$
& $0.19 \pm 0.04$ \\
\textbf{StepKV (Ours)}
& \cellcolor{BudgetBlue!20}20\% & $\textbf{18.07} \pm 0.81$ & $\textbf{24.29} \pm 1.70$
& $\textbf{13.47} \pm 1.79$ & $\textbf{17.82} \pm 0.97$
& $\textbf{2.80} \pm 0.53$ & $\textbf{8.10} \pm 1.34$ \\
\midrule
\rowcolor{gray!15}
\multicolumn{8}{c}{\textbf{Llama3.1-8B-Instruct}} \\
Full KV
& 100\% & $25.27 \pm 0.23$ & $34.21 \pm 2.06$
& $22.20 \pm 0.35$ & $25.01 \pm 0.39$
& $4.40 \pm 1.25$ & $7.74 \pm 1.41$ \\
ReAct
& -- & $6.80 \pm 0.00$ & $9.67 \pm 0.19$
& $5.00 \pm 1.22$ & $5.74 \pm 1.04$
& $1.27 \pm 0.42$ & $1.91 \pm 0.38$ \\
\cmidrule{1-8}
H$_2$O
& \cellcolor{BudgetBlue!10}50\% & $11.80 \pm 0.92$ & $16.51 \pm 1.94$
& $8.60 \pm 0.20$ & $10.97 \pm 0.94$
& $3.10 \pm 0.62$ & $7.22 \pm 0.41$ \\
TOVA
& \cellcolor{BudgetBlue!10}50\% & $13.33 \pm 0.23$ & $18.58 \pm 1.02$
& $9.27 \pm 0.42$ & $11.35 \pm 0.87$
& $2.50 \pm 0.95$ & $6.49 \pm 0.36$ \\
TokenSkipping
& \cellcolor{BudgetBlue!10}50\%
& $6.80 \pm 2.28$
& $10.00 \pm 1.86$
& $4.07 \pm 0.09$
& $5.01 \pm 0.13$
& $1.07 \pm 0.25$
& $4.29 \pm 0.02$ \\
\textbf{StepKV (Ours)}
& \cellcolor{BudgetBlue!10}50\% & $\textbf{25.07} \pm 0.92$ & $\textbf{33.52} \pm 3.33$
& $\textbf{22.40} \pm 0.92$ & $\textbf{25.43} \pm 1.32$
& $\textbf{5.43} \pm 0.85$ & $\textbf{10.52} \pm 1.57$ \\
\cmidrule{1-8}
H$_2$O
& \cellcolor{BudgetBlue!20}20\% & $0.53 \pm 0.46$ & $1.08 \pm 0.46$
& $0.87 \pm 0.46$ & $1.18 \pm 0.76$
& $0.13 \pm 0.12$ & $0.36 \pm 0.19$ \\
TOVA
& \cellcolor{BudgetBlue!20}20\% & $0.60 \pm 0.00$ & $1.25 \pm 0.10$
& $0.60 \pm 0.20$ & $0.71 \pm 0.24$
& $0.00 \pm 0.00$ & $0.20 \pm 0.07$ \\
TokenSkipping
& \cellcolor{BudgetBlue!20}20\%
& $0.07 \pm 0.01$
& $0.07 \pm 0.01$
& $0.00 \pm 0.00$
& $0.00 \pm 0.00$
& $0.00 \pm 0.00$
& $0.00 \pm 0.00$ \\
\textbf{StepKV (Ours)}
& \cellcolor{BudgetBlue!20}20\% & $\textbf{19.93} \pm 0.12$ & $\textbf{25.46} \pm 1.04$
& $\textbf{13.80} \pm 3.22$ & $\textbf{15.21} \pm 3.57$
& $\textbf{3.03} \pm 1.21$ & $\textbf{5.71} \pm 1.65$ \\
\bottomrule
\end{tabular*}
\label{tab:main_results}
\end{table*}
\vspace{-0.2cm}
\subsection{Experimental Setup}
\label{sec:exp_setup}

\noindent\textbf{Datasets.}
We evaluate StepKV on four representative agent reasoning benchmarks, including HotpotQA~\cite{hotpotqa}, 2WikiMultihopQA~\cite{2wiki}, MuSiQue~\cite{musique}, and BrowseComp-Plus~\cite{browsecomp-plus}. HotpotQA, 2WikiMultihopQA, and MuSiQue are used to evaluate basic multi-hop agent question answering, where performance is measured by Exact Match (EM) and token-level F1. BrowseComp-Plus is a challenging long-horizon web reasoning benchmark whose reasoning trajectories can exceed 100K tokens. On this benchmark, we additionally report inference latency and the final KV cache size to evaluate efficiency under extremely long-context reasoning.

\noindent\textbf{Baselines.}
We compare StepKV with three representative KV cache compression methods: H$_2$O~\cite{zhang2023h2o}, TOVA~\cite{tova}, and TokenSkipping~\cite{tokenskipping}, together with the Full KV and ReAct baselines. Specifically, we categorize existing decoding-stage KV cache management approaches into three representative paradigms. H$_2$O represents \emph{accumulated score-based} methods, which retain tokens according to accumulated attention statistics; TOVA represents \emph{online attention-based} methods, which perform dynamic eviction based on attention scores during decoding; and TokenSkipping represents \emph{heuristic-based} methods, which reduce cache usage through predefined token-skipping rules. Together, these methods cover three representative paradigms of existing KV cache management during decoding.

\noindent\textbf{Implementation details} are provided in Appendix~\ref{implementation}.

\vspace{-0.2cm}
\subsection{Performance under Tight KV Budgets (RQ1)}
\label{sec:rq1}
\textbf{StepKV preserves multi-hop QA accuracy much more reliably than token-level pruning under constrained cache budgets.} Table~\ref{tab:main_results} compares Full KV and ReAct reference settings with H$_2$O, TOVA, TokenSkipping and StepKV across three agent multi-hop QA datasets and two backbone models, Qwen2.5-7B-Instruct~\cite{qwen2.5} and Llama-3.1-8B-Instruct~\cite{grattafiori2024llama}.

\textbf{The improvement is consistent across models, datasets, and metrics.} In Table~\ref{tab:main_results}, StepKV is the best compressed method for every model-budget-dataset combination, as indicated by the bolded EM and F1 values. This pattern holds on both backbones and all three multi-hop QA datasets, suggesting that the gain is not tied to a single model or benchmark.

\textbf{The advantage becomes largest when the KV budget is most restrictive.} At the 20\% budget in Table~\ref{tab:main_results}, H$_2$O, TOVA, and TokenSkipping frequently degrade to near-zero EM and very low F1, especially on HotpotQA and 2WikiMultihopQA. Through further analysis, we attribute this severe performance degradation to \textbf{\emph{Reasoning Continuity Disruption}}. When critical intermediate information is aggressively removed, the agent loses awareness of previously generated reasoning context and may repeatedly generate similar tokens or redundant reasoning patterns, preventing effective progress toward the final answer.

In contrast, StepKV consistently maintains usable performance across all datasets. We attribute this robustness to two key designs. First, StepKV adopts a \textbf{step-wise generation before compression} strategy, which allows the agent to complete an entire reasoning step before performing KV eviction. Unlike token-only compression methods that prune tokens within an ongoing reasoning step, this design preserves the internal coherence and information flow of the current reasoning process. Second, StepKV introduces a \textbf{step-aware scoring mechanism} to estimate the importance of each reasoning step and combines it with token-level importance for cache selection. By assigning higher retention priority to tokens belonging to critical reasoning steps, StepKV preserves essential reasoning trajectories while effectively removing redundant information.

\begin{figure}[t]
    \centering
    \includegraphics[width=0.48\linewidth]{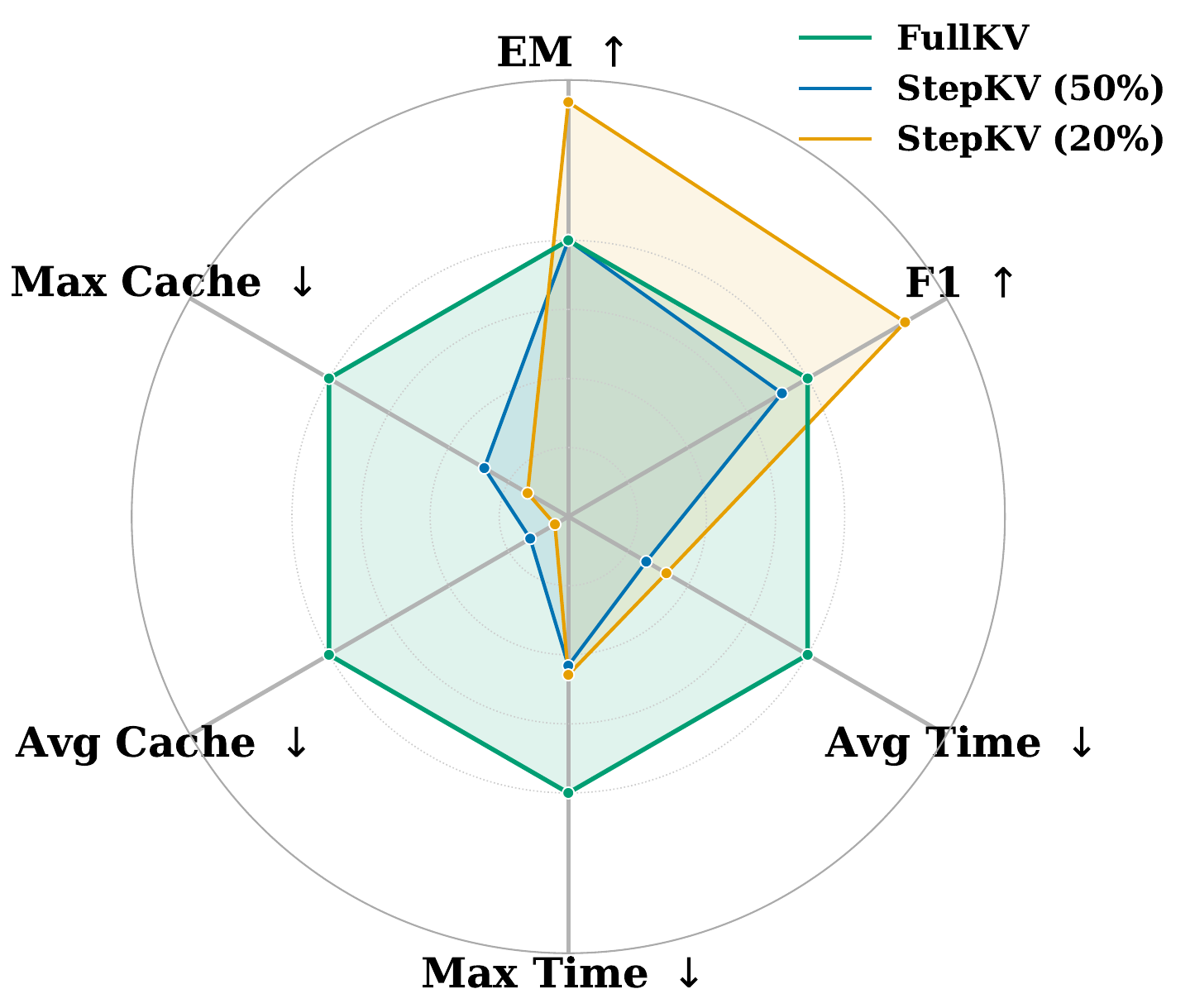}\hfill
    \includegraphics[width=0.48\linewidth]{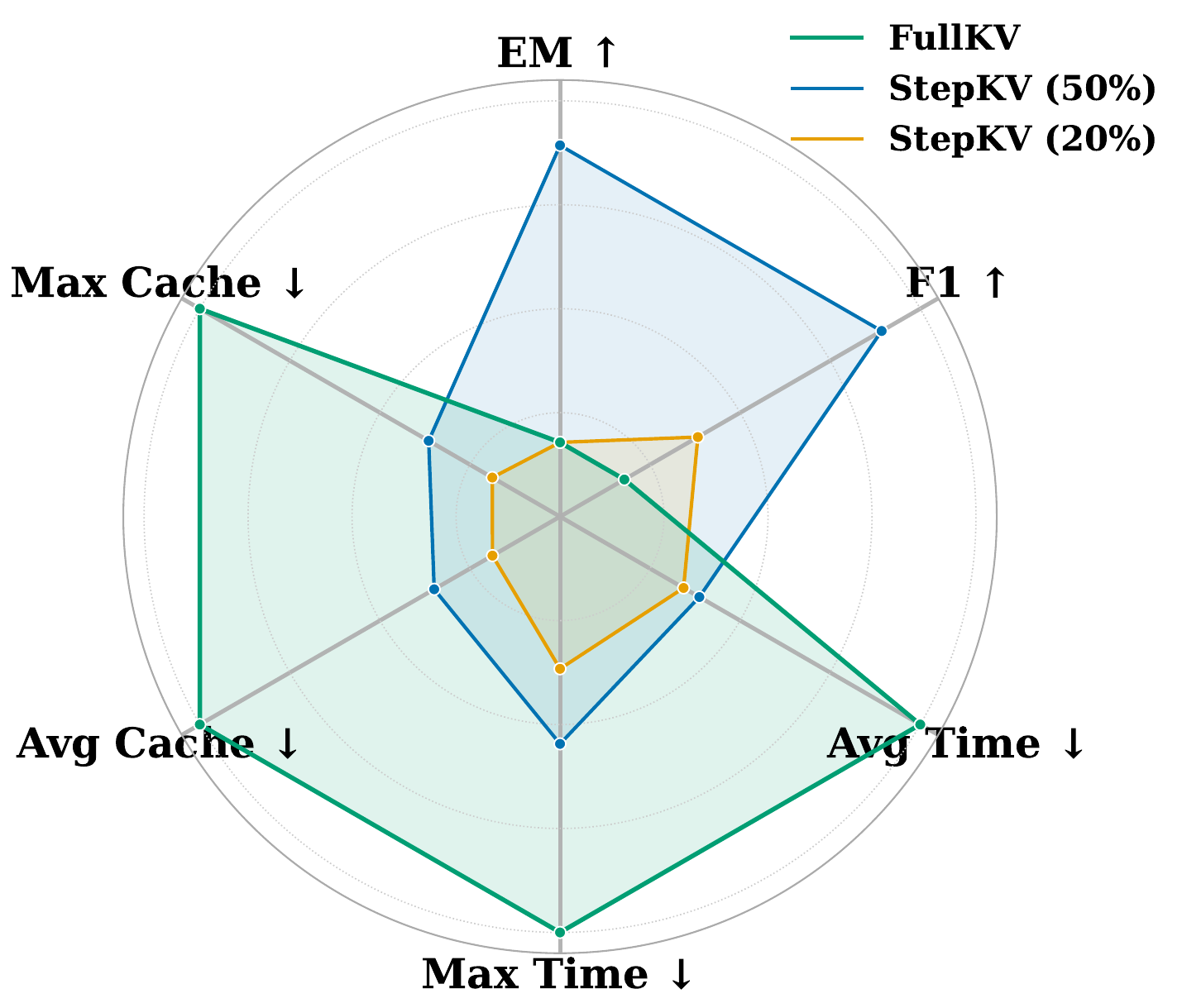}
    \caption{Performance--cost trade-off on BrowseComp-Plus for Qwen2.5-7B-Instruct (left) and Llama3.1-8B-Instruct (right). Lower normalized cache and time values indicate lower cost.}
    \label{fig:browsecomp-plus}
     \vspace{-0.5cm}
\end{figure}
\textbf{Moderate compression can preserve much of the full-cache behavior when allocation is step-aware.} At the 50\% budget, StepKV substantially narrows the gap to Full KV and sometimes approaches the reference results, whereas Token-only methods still lose large portions of the QA accuracy. This shows that the benefit is not merely a rescue effect at extreme budgets; it also improves the practical operating range of cache compression.

\FloatBarrier

\begin{figure*}[h]
    \centering
    \includegraphics[width=0.9\linewidth]{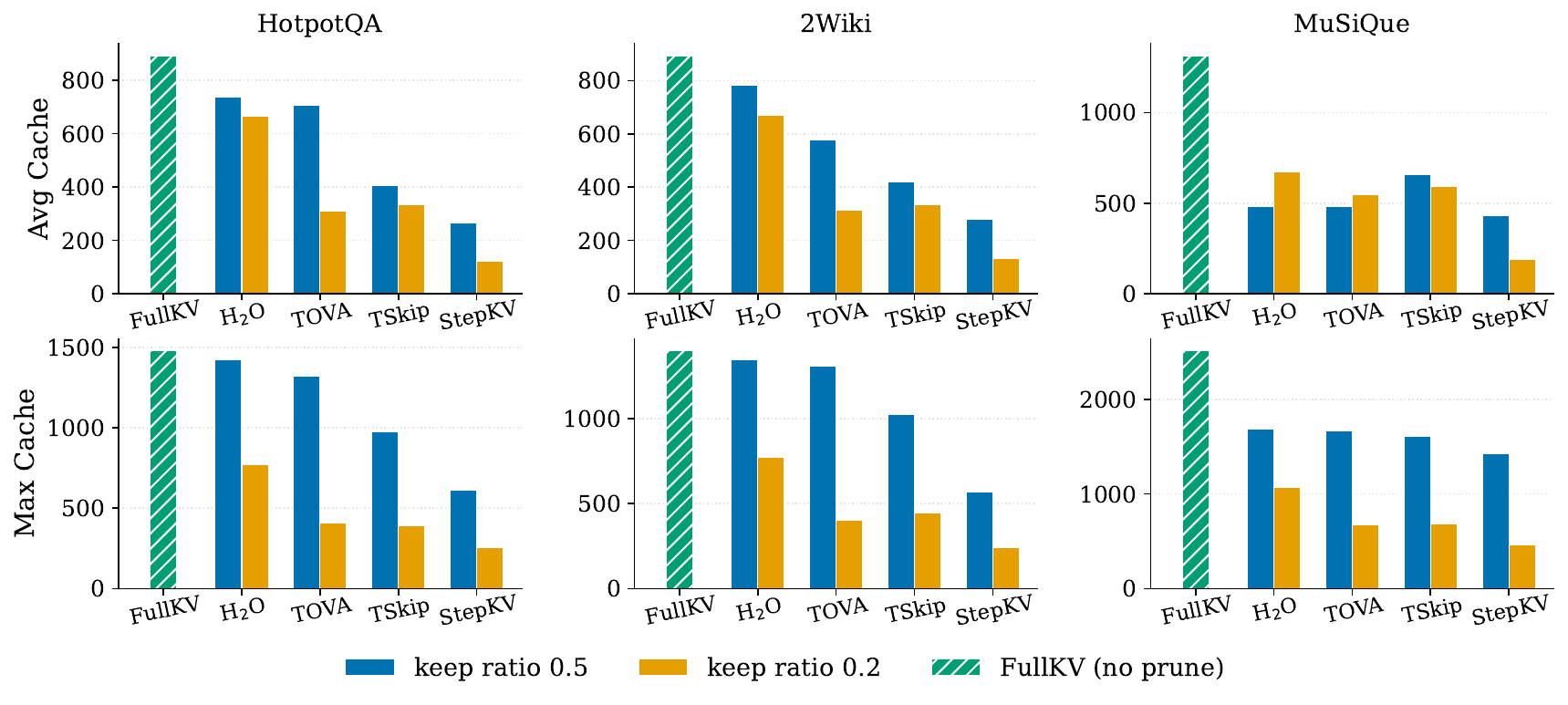}
    \caption{Comparison of inference latency and final KV cache size under different KV cache strategies.}
    \label{fig:latency_kv}
     \vspace{-0.3cm}
\end{figure*}
\vspace{0.4cm}
\textbf{The advantage extends to substantially more challenging and long-horizon agent trajectories.}
Figure~\ref{fig:browsecomp-plus} evaluates StepKV on BrowseComp-Plus, a
challenging long-horizon web reasoning benchmark with interaction histories
exceeding 100K tokens. Such extremely long trajectories pose significant
challenges for KV cache management, as redundant historical context may
interfere with subsequent reasoning. As shown in Figure~\ref{fig:browsecomp-plus},
StepKV substantially reduces KV cache size and inference latency while even
improving EM and F1 over Full KV under compressed settings. This demonstrates
that ultra-long agent trajectories contain redundant information and that
StepKV can effectively identify low-utility tokens while preserving critical
reasoning dependencies. The significant latency reduction further verifies the
scalability of StepKV for long-context agent inference.

\vspace{-0.2cm}
\subsection{Practical Cost in general Agent Reasoning (RQ2)}
\label{sec:rq2}
\textbf{StepKV turns cache reduction into a usable cost-quality trade-off rather than a pure memory-saving heuristic.}
Figure~\ref{fig:latency_kv} compares peak KV cache statistics and inference latency across Full KV, token-only methods, and StepKV under the same budget settings, while Table~\ref{tab:main_results} provides the corresponding task-quality evidence for compressed QA inference.

\textbf{Cache savings are meaningful only when reasoning continuity is preserved.}
As shown in Figure~\ref{fig:latency_kv}, compressed settings generally reduce KV cache size compared with Full KV. However, token-only methods may fail to achieve the expected reduction under the same compression ratio. Aggressive token pruning can disrupt the reasoning trajectory, causing the agent to lose critical intermediate information and generate longer, redundant reasoning sequences. As a result, the final KV cache can exceed the expected budget despite applying the same compression ratio.

\textbf{StepKV effectively avoids this reasoning overhead by preserving reasoning continuity.}
Unlike token-only compression methods that perform eviction directly at the token level during generation, StepKV preserves the integrity of each reasoning step and prioritizes tokens from important reasoning steps through step-aware scoring. As a result, StepKV achieves substantially lower inference latency and smaller KV cache usage while maintaining higher EM and F1 under the same budgets. These results demonstrate that effective KV cache compression requires not only reducing stored tokens, but also preserving the critical reasoning trajectory, enabling StepKV to approach Full KV performance with significantly fewer cached tokens.
\begin{table}[h]
\centering
\footnotesize
\caption{Detailed component ablation of Step Score in StepKV on MuSiQue.}
\label{tab:stepkv-ablation-musique}

\resizebox{\columnwidth}{!}{
\begin{tabular}{lcccc}
\toprule
& \multicolumn{2}{c}{Qwen2.5-7B-Instruct}
& \multicolumn{2}{c}{Llama-3.1-8B-Instruct}
\\
\cmidrule(lr){2-3}
\cmidrule(lr){4-5}
Variant
& EM & F1
& EM & F1
\\
\midrule
\textbf{Full StepKV}
& 7.00 & 12.49
& 4.80 & 8.93
\\
w/o redundancy
& -2.40 & -5.52
& -2.00 & -4.60
\\
w/o evidence gain
& -3.40 & -5.47
& -0.80 & -3.25
\\
w/o validity
& -2.80 & -5.58
& -2.00 & -4.79
\\
w/o reuse update
& -3.00 & -6.45
& -1.40 & -3.73
\\
\bottomrule
\end{tabular}
}

\end{table}

\begin{table}[h]
\centering
\footnotesize
\setlength{\tabcolsep}{4pt}
\caption{Component ablation of StepKV on three benchmarks.}
\label{tab:stepkv-ablation-two-models}
\begin{tabular}{lcccccc}
\toprule[1.2pt]
\multicolumn{7}{c}{\textbf{Qwen2.5-7B-Instruct}}\\
\midrule
Variant
& \multicolumn{2}{c}{HotpotQA}
& \multicolumn{2}{c}{2Wiki}
& \multicolumn{2}{c}{MuSiQue}
\\
& EM & F1
& EM & F1
& EM & F1
\\
\midrule
\textbf{Full StepKV}
& 21.60 & 29.96
& 23.60 & 26.85
& 7.00 & 12.49
\\
w/o token score
& +0.20 & -2.33
& +1.40 & +0.27
& -4.60 & -8.43
\\
w/o step score
& -5.60 & -7.30
& -15.60 & -12.45
& -3.40 & -5.77
\\
\midrule[0.8pt]
\multicolumn{7}{c}{\textbf{Llama-3.1-8B-Instruct}}\\
\midrule
Variant
& \multicolumn{2}{c}{HotpotQA}
& \multicolumn{2}{c}{2Wiki}
& \multicolumn{2}{c}{MuSiQue}
\\
& EM & F1
& EM & F1
& EM & F1
\\
\midrule
\textbf{Full StepKV}
& 25.60 & 31.60
& 21.40 & 23.90
& 4.80 & 8.93
\\
w/o token score
& -2.00 & -2.58
& -0.80 & -1.16
& -2.60 & -4.87
\\
w/o step score
& -4.80 & -5.16
& -9.00 & -9.11
& -2.20 & -3.83
\\
\bottomrule[1.2pt]
\end{tabular}
\end{table}

\subsection{Reasoning Depth and Compressed-Cache Performance (RQ3)}
\label{sec:rq3}
\begin{figure}[h]
    \centering
    \includegraphics[width=\linewidth]{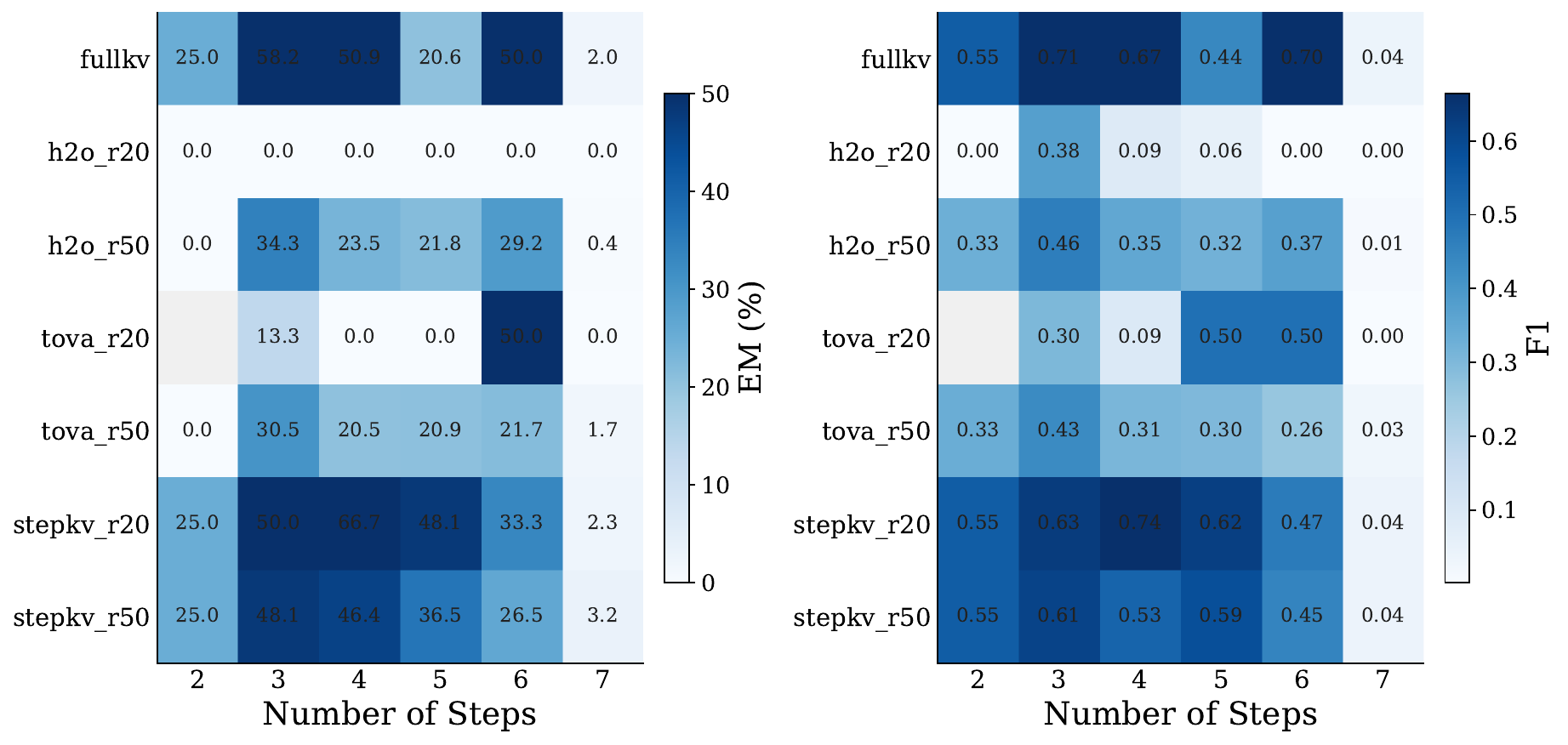}
    \caption{Accuracy grouped by the number of reasoning steps.}
    \label{fig:stepwise_accuracy}
    \vspace{-0.5cm}
\end{figure}

\textbf{Compression becomes harder as reasoning trajectories grow deeper.} Figure~\ref{fig:stepwise_accuracy} groups samples by reasoning-step count; deeper bins show lower EM/F1, especially under tighter budgets.

\textbf{The depth effect reflects cross-step dependence rather than only longer token sequences.} In Figure~\ref{fig:stepwise_accuracy}, deeper samples require earlier evidence across more later decisions, matching the continuity-disruption analysis in Section~\ref{sec:motivation}.

\vspace{0.2cm}

\begin{figure}[h]
    \centering
    \includegraphics[width=\linewidth]{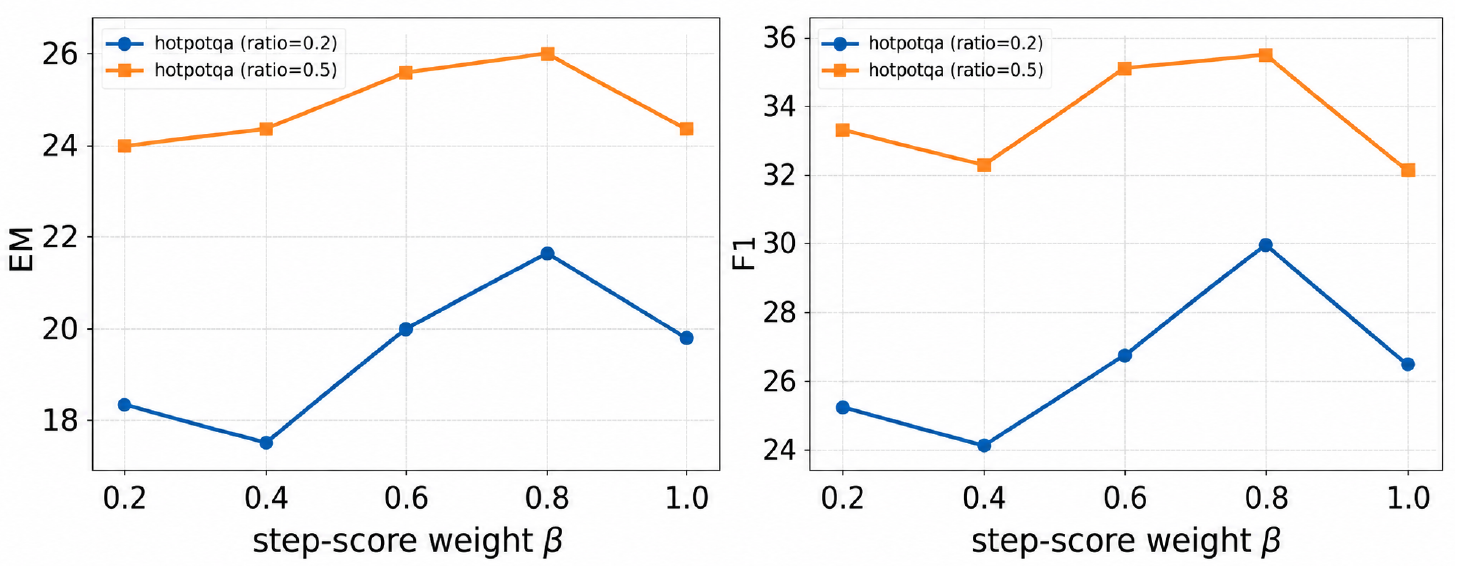}
    \caption{Influence of the step score hyperparameter on model performance.}
    \label{fig:hyper_parameter}
    \vspace{-0.5cm}
\end{figure}

\subsection{Contribution of Step-Level Signals (RQ4)}
\label{sec:rq4}

\textbf{Step-level utility is most effective when paired with token saliency.} Figure~\ref{fig:hyper_parameter} varies the relative weight of $S_k$ in the weighted token-step ranking; intermediate values perform best.

\begin{figure}[t]
    \centering
    \includegraphics[width=\linewidth]{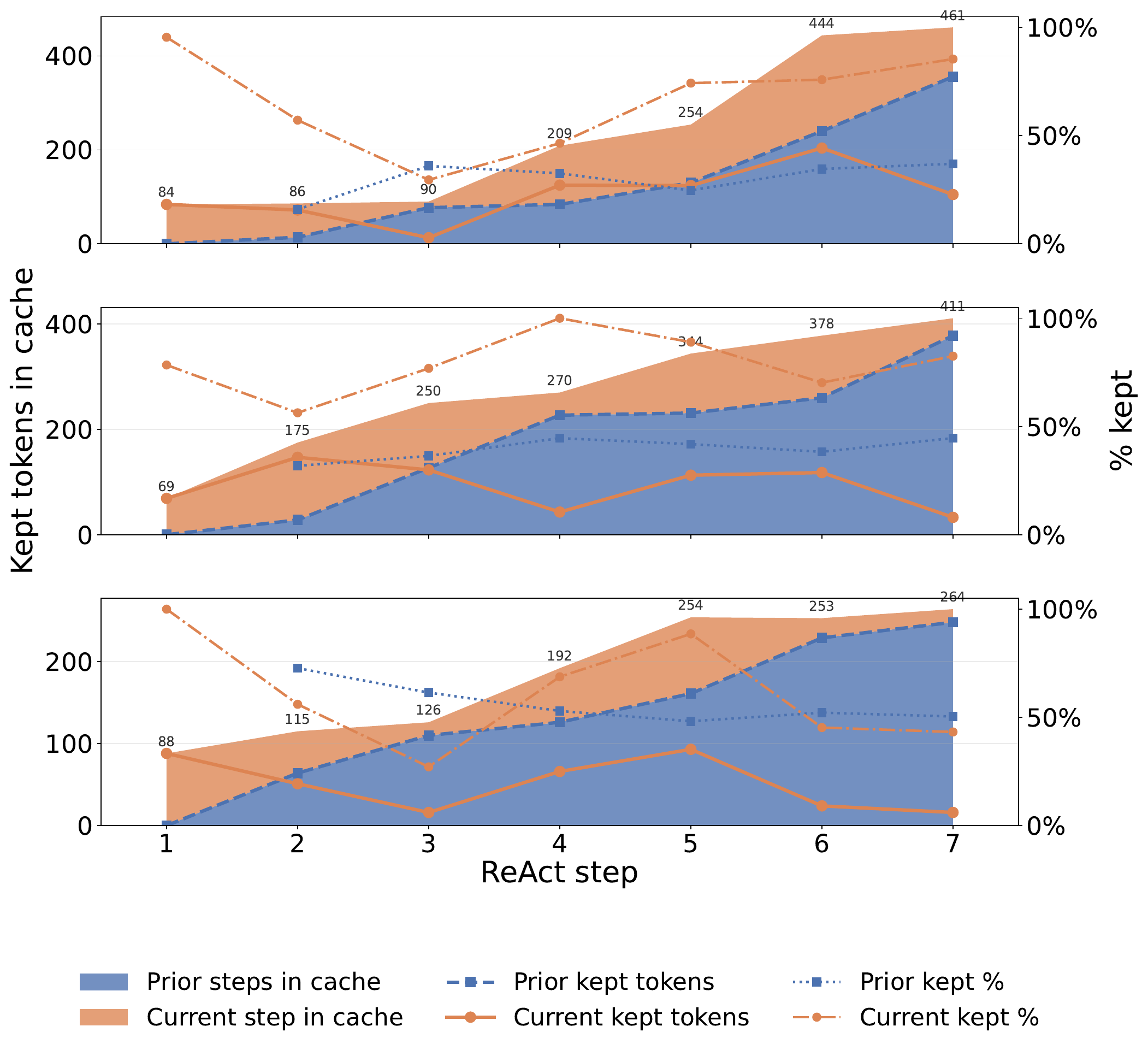}
    \caption{Retained-cache composition across reasoning steps for H$_2$O (top), TOVA (middle), and StepKV (bottom). Filled regions separate prior-step and current-step cache populations; marker lines report kept-token counts, and dotted or dash-dotted lines report the corresponding retained percentages.}
    \label{fig:cache-composition}
    \vspace{-0.6cm}
\end{figure}

\textbf{StepKV maintains a more balanced cache across prior and newly completed steps.} Figure~\ref{fig:cache-composition} decomposes the retained cache at each reasoning boundary into tokens from prior steps and tokens from the current step. Under H$_2$O and TOVA, the retained fraction of prior-step tokens drops sharply or fluctuates as token-wise saliency reallocates the budget. StepKV instead maintains a larger and more stable share of prior-step context through the early and middle stages of the trajectory, while continuing to retain tokens from the newly completed step. The balance still changes across boundaries, indicating utility-sensitive allocation rather than uniform protection of every step.

\textbf{This allocation directly addresses the two continuity failures observed in Figure~\ref{fig:step_analysis}.}
In Figure~\ref{fig:step_analysis}(\subref{fig:cohort_step_token}), StepKV preserves early step cohorts across later reasoning stages, while H$_2$O and TOVA rapidly discard them. Meanwhile, the nonzero current-step allocation in Figure~\ref{fig:cache-composition} prevents historical retention from overwhelming recent context. As shown in Figure~\ref{fig:step_analysis}(\subref{fig:step_drop}), StepKV performs more selective eviction across source steps instead of fragmenting the reasoning trajectory. These results demonstrate that step utility maintains coherent step context, delayed reuse restores the importance of previously useful steps, and token saliency provides fine-grained selection within each step.

\textbf{Step utility and token saliency provide complementary evidence for cache selection.} Table~\ref{tab:stepkv-ablation-two-models} compares the full method with variants that remove either signal. Removing the step score causes the larger and more consistent degradation across models and benchmarks, while removing token saliency also hurts most settings, especially MuSiQue. The full method therefore benefits from modeling trajectory-level importance without discarding token-level evidence.

\textbf{Each field-level signal makes a distinct contribution to the step score.} Table~\ref{tab:stepkv-ablation-musique} shows that removing observation validity, evidence gain, action redundancy, or delayed reuse reduces both EM and F1 on MuSiQue. No single cue accounts for the full improvement, supporting the use of immediate step quality together with relevance revealed later in the trajectory.

\section{Conclusion}
We propose \textbf{StepKV}, a step-aware KV cache compression framework tailored for LLM agents. Traditional token-centric methods often degrade agent reasoning by ignoring its structured nature and dropping crucial context prematurely. To address this, StepKV introduces a \textbf{step-level perspective} that effectively captures the structured reasoning trajectories of agents and the delayed reuse of intermediate steps. By \textbf{jointly optimizing step-level utility and token-level saliency}, StepKV preserves essential reasoning paths while drastically reducing KV cache memory and computational overhead. Extensive experiments show that StepKV achieves superior performance and a markedly better accuracy--efficiency trade-off across complex tool-use multi-hop QA and long-horizon web search tasks.

\bibliographystyle{ACM-Reference-Format}
\bibliography{stepkv_refs}
\clearpage

\appendix
\raggedbottom

\section{Notation Summary}
\label{app:notation}

\begin{table}[H]
\caption{Notation used in the problem formulation and methodology.}
\label{tab:notation}
\centering
\footnotesize
\setlength{\tabcolsep}{3pt}
\renewcommand{\arraystretch}{1.08}
\begin{tabular}{@{}>{\raggedright\arraybackslash}p{0.27\columnwidth}>{\raggedright\arraybackslash}p{0.67\columnwidth}@{}}
\toprule
Notation & Description \\
\midrule
\rowcolor{gray!12}\multicolumn{2}{l}{\textbf{Agent trajectory}} \\
$q$ & User query. \\
$t$ & Index of the latest completed step and the corresponding pruning boundary. \\
$k,u$ & Reasoning-step indices; $u$ typically indexes an earlier step. \\
$\mathrm{Step}_k$ & The $k$-th completed agent interaction step. \\
$\mathrm{Th}_k,\mathrm{Act}_k,\mathrm{Obs}_k$ & Thought, Action, and Observation fields of step $k$. \\
$\mathcal{S}^{(t)}$ & Indices of the finalized steps after step $t$. \\
$\tau^{(t)}$ & Text trajectory containing $q$ and all finalized steps through $t$. \\
\rowcolor{gray!12}\multicolumn{2}{l}{\textbf{KV cache and budget}} \\
$i$ & Index of a prunable trajectory token. \\
$\mathcal{I}^{(t)},N_t$ & Prunable token-index set after step $t$ and its size, respectively. \\
$\mathcal{C}^{(t)}$ & Full KV cache at the pruning boundary after step $t$. \\
$\mathcal{C}_{\mathrm{prot}},\mathcal{C}_{\mathrm{traj}}^{(t)}$ & Protected prompt/query cache and prunable trajectory cache. \\
$\rho,B_t$ & KV keep ratio and resulting token budget $\max(1,\lfloor\rho N_t\rfloor)$. \\
$\mathcal{K}^{(t)}$ & Indices of the trajectory tokens retained under budget $B_t$. \\
$\widehat{\mathcal{C}}^{(t)}$ & Compressed KV cache formed from protected and retained entries. \\
$\|$ & Concatenation of text fields or cache segments. \\
\rowcolor{gray!12}\multicolumn{2}{l}{\textbf{Step and token scoring}} \\
$\operatorname{sim}(a,b)$ & Overlap-based textual similarity between fields $a$ and $b$. \\
$\mathrm{valid}_k$ & Indicator that $\mathrm{Obs}_k$ contains a usable result. \\
$\mathrm{gain}_k$ & Evidence introduced by step $k$ beyond earlier steps. \\
$\mathrm{red}_k$ & Redundancy of $\mathrm{Act}_k$ with earlier actions. \\
$r_k$ & Completion score assigned when step $k$ is finalized. \\
$c_k$ & Reuse accumulator recording delayed relevance of step $k$. \\
$S_k$ & Step utility combining completion and accumulated reuse. \\
$T_i$ & Conventional attention-based saliency of token $i$. \\
$P_i$ & Retention priority of token $i$, combining $T_i$ with its source-step utility. \\
$\lambda,\eta$ & Action-redundancy penalty and reuse-update scale. \\
$w_r,w_c$ & Weights of the completion and reuse terms in $S_k$. \\
$r_{\min},r_{\max},S_{\max}$ & Clipping bounds for the completion score and step utility. \\
$\alpha,\beta$ & Token-saliency and step-utility weights in the final ranking. \\
\bottomrule
\end{tabular}
\end{table}
\FloatBarrier

\section{Related Work}
\label{app:relatedworks}

\subsection{LLM Agents}

LLM agents extend language-model inference from single-pass generation to
interactive trajectories. Agent frameworks such as ReAct interleave reasoning
and tool use through repeated Thought, Action, and Observation steps
~\cite{yao2023react}, while related systems leverage external tools, web
interaction, reflection, or self-improvement mechanisms to enhance task
completion over multiple interactions
~\cite{schick2023toolformer,nakano2022webgpt,shinn2023reflexion}.

Beyond tool use and planning, recent studies have emphasized the importance of
long-term context management and memory mechanisms for agent systems.
MemGPT~\cite{memgpt} views LLMs as systems with hierarchical memory, where
information is selectively managed across different memory tiers.
Generative Agents~\cite{park2023generative} and MemoryBank
~\cite{zhong2024memorybank} further explore how agents maintain, retrieve, and
reuse historical experiences during extended interactions. These works
highlight an important property of agent trajectories: the value of
intermediate information is not always observable when it is generated, and
may only emerge when later interactions depend on previously acquired
knowledge.

This setting changes the role of context in LLM inference. Rather than
conditioning only on a fixed prompt and a generated answer prefix, an agent
continuously accumulates intermediate observations, retrieved evidence, and
decisions whose usefulness may appear at different stages of reasoning. Most
agent research focuses on planning, tool selection, retrieval, memory design,
and task performance, while the KV cache is often treated as an underlying
inference mechanism. Recent work has started to investigate cache management
for agentic inference, including model-driven, directive-based, and
intent-aware approaches that determine which parts of an interaction
trajectory should be retained
~\cite{kariyappa2026sidequest,ma2026leyline,li2026intentkv}.

StepKV is complementary to these directions. Instead of designing an
additional external memory module or changing the agent framework, StepKV
focuses on fine-grained retention within the KV cache and allocates a fixed
cache budget using both token-level saliency and step-level utility. This
distinction is important because agent trajectories contain semantic step
boundaries that are not explicitly represented in a flat token stream, and
preserving information aligned with these boundaries can help maintain
reasoning continuity under aggressive cache compression.

\subsection{KV Cache Management}

A broad line of research seeks to reduce the memory, bandwidth, and
computational costs of long-context LLM inference through architectural
changes, context compression, selective attention, and explicit KV cache
management.

\textbf{Architecture- and attention-level optimization.}
Architecture-level approaches reduce the cost of representing or processing
long contexts. Grouped-Query Attention (GQA)~\cite{ainslie2023gqa} and
Multi-Head Latent Attention (MLA)~\cite{deepseekai2024deepseekv2} reduce the
memory footprint of KV representations by modifying the attention
architecture, while Native Sparse Attention~\cite{yuan2025nativesparse}
reduces long-context computation by selectively activating relevant tokens.
State-space and hybrid architectures, such as Mamba~\cite{gu2024mamba} and
Gated Delta Networks~\cite{yang2025gateddelta}, further reduce reliance on
full attention for modeling long-range dependencies. Beyond architectural
optimization, adaptive computation methods dynamically allocate resources
according to input characteristics. Adaptive Attention Span
~\cite{sukhbaatar2019adaptive} learns different attention ranges for different
tokens, while Mixture-of-Depths~\cite{raposo2024mixture} dynamically allocates
computation across tokens during inference. Although these approaches improve
long-context efficiency, they mainly optimize how contextual representations
are computed or stored and do not explicitly determine which historical KV
entries should be retained as an agent trajectory evolves.

\textbf{Long-context compression.}
Another line of work focuses on removing redundant contextual information
before or during inference. LLMLingua~\cite{jiang2023llmlingua} and
LongLLMLingua~\cite{jiang2024longllmlingua} compress input contexts by
identifying less informative tokens, improving efficiency in long-context
scenarios. Meanwhile, Lost in the Middle~\cite{liu2023lost} demonstrates that
LLMs may struggle to effectively utilize information distributed across long
contexts, suggesting that retaining all historical information does not
necessarily lead to better reasoning. These studies motivate selective context
retention strategies that preserve useful information while reducing
unnecessary context.

\textbf{Decode-stage KV cache management.}
In contrast to input compression, decode-stage KV cache management selectively
retains historical KV states as the cache grows during autoregressive
generation. Existing approaches commonly estimate the importance of individual
tokens using several types of signals.

First, \textbf{accumulated-importance methods} use historical statistics
collected throughout generation. H$_2$O~\cite{zhang2023h2o} identifies
heavy-hitter tokens according to accumulated attention scores, while
Scissorhands~\cite{scissorhands} studies persistent token importance across
generation steps. These methods effectively capture long-term relevance but
primarily rely on token-level historical statistics.

Second, \textbf{recent-attention methods} estimate cache importance from
current or recent attention distributions. TOVA~\cite{tova} performs online
eviction by retaining tokens that receive higher attention from recent
queries; SnapKV~\cite{li2024snapkv} analyzes local attention patterns and
preserves representative tokens from important attention regions; and
Quest~\cite{tang2024quest} improves cache selection through query-aware
importance estimation. However, attention patterns at the current decoding
stage may not fully reflect the future contribution of tokens in long-horizon
reasoning.

Third, \textbf{heuristic and reasoning-aware methods} reduce KV cache size
using predefined rules, structural assumptions, or additional reasoning
signals. TokenSkipping~\cite{tokenskipping} periodically removes tokens using
heuristic skipping strategies, whereas StreamingLLM~\cite{xiao2024streamingllm}
preserves attention-sink tokens and recent context based on observed attention
behavior. Recent reasoning-oriented methods, including R-KV
~\cite{cai2026rkv}, RaaS~\cite{hu2025raas}, LazyEviction
~\cite{zhang2025lazyeviction}, ThinKV~\cite{ramachandran2026thinkv}, and
SideQuest~\cite{kariyappa2026sidequest}, further explore
redundancy-aware or reasoning-aware cache reduction.

Nevertheless, most existing methods remain predominantly token-centric or do
not explicitly model the semantic utility of complete reasoning steps. In
contrast, StepKV introduces step-aware cache allocation for agent reasoning:
rather than treating all tokens independently, it combines step-level utility
with token-level saliency to preserve critical reasoning trajectories while
removing redundant context.

\section{Analysis of Step-aware Score Scaling}
\label{step_score_analysis}

\subsection{Step Score}
\paragraph{Definition of step utility.}
Let $\mathcal{S}$ denote the set of reasoning steps generated during agent inference. 
Instead of assigning importance solely from individual tokens, StepKV introduces a step-level utility function:

\begin{equation}
U:\mathcal{S}\rightarrow[0,U_{\max}],
\end{equation}

where $U_s$ represents the contribution of reasoning step $s$ to the overall reasoning trajectory.

The step utility is computed based on two complementary signals: the immediate progression signal and the delayed reuse signal:

\begin{equation}
U_s=
\operatorname{clip}
\left(
w_r R_s+w_c\log(1+C_s),
0,U_{\max}
\right),
\end{equation}

where $w_r,w_c\geq0$ are weighting coefficients.

The progression signal is defined as:

\begin{equation}
R_s=
\mathrm{succ}_s+
\mathrm{nov}_s-
\lambda\mathrm{rep}_s ,
\end{equation}

where $\mathrm{succ}_s\in\{0,1\}$ indicates whether the step produces valid progress,
$\mathrm{nov}_s\in[0,1]$ measures the novelty ratio of the step, and
$\mathrm{rep}_s\geq0$ denotes the repetition degree.

Since the maximum positive contribution of the progression term is bounded by:

\begin{equation}
\mathrm{succ}_s+\mathrm{nov}_s\leq2,
\end{equation}

the progression signal satisfies:

\begin{equation}
R_s\in(-\infty,2].
\end{equation}

To capture whether information from an earlier step is reused by later reasoning steps, we accumulate the delayed reuse signal:

\begin{equation}
C_s^{(t)}
=
C_s^{(t-1)}
+
\Delta C_{s\rightarrow t},
\end{equation}

where:

\begin{equation}
0\leq\Delta C_{s\rightarrow t}\leq1 .
\end{equation}

Therefore:

\begin{equation}
C_s\geq0.
\end{equation}

The logarithmic transformation prevents long reasoning trajectories from causing unbounded growth of the reuse signal. After applying clipping, the final step utility is guaranteed to satisfy:

\begin{equation}
U_s\in[0,U_{\max}].
\end{equation}

\paragraph{Combination with token-level importance.}
For each token $i\in\mathcal{T}$, let $s(i)$ denote its corresponding reasoning step. 
StepKV combines token-level importance and step-level utility as:

\begin{equation}
\mathrm{Score}_i
=
\alpha H_i+
\beta U_{s(i)},
\end{equation}

where $H_i$ represents the token-level importance score obtained from attention statistics, and $\alpha,\beta\geq0$ control the contribution of each component.

Assuming the token importance is normalized as:

\begin{equation}
H_i\in[0,H_{\max}],
\end{equation}

the token score has the following range:

\begin{equation}
\mathrm{Score}_i
\in
[0,\alpha H_{\max}+\beta U_{\max}].
\end{equation}

\paragraph{Rationale of score range design.}
The step utility is designed as a bounded auxiliary signal rather than a replacement for token-level importance. 
The objective is to introduce reasoning-level awareness while preserving the fine-grained discrimination ability of token scores.

Specifically, the contribution of the step utility is bounded by:

\begin{equation}
0\leq
\beta U_{s(i)}
\leq
\beta U_{\max}.
\end{equation}

Therefore, the step-level component mainly affects tokens with similar token-level importance scores. 
For two tokens $i$ and $j$, the token from a more important reasoning step is preferred when:

\begin{equation}
\alpha(H_i-H_j)
<
\beta(U_{s(j)}-U_{s(i)}).
\end{equation}

However, when a token has substantially higher intrinsic importance, the token-level score can overcome the step-level preference:

\begin{equation}
\alpha(H_i-H_j)
>
\beta(U_{s(j)}-U_{s(i)}).
\end{equation}

This property prevents the step utility from dominating cache selection. 
Instead, StepKV follows a hierarchical importance principle: the step utility provides global reasoning-level guidance by protecting valuable reasoning steps, while the token-level score preserves critical tokens within or across different steps.

Consequently, StepKV can retain important reasoning trajectories without sacrificing the ability to select highly salient individual tokens.

\subsection{{\color{black}Time Complexity Analysis}}
At the pruning boundary after step~$t$, new-step scoring and previous-step updating compare the current fields with at most $t-1$ earlier steps. Because each field uses a fixed cap of 64 evidence cues, each overlap computation is bounded, yielding $O(t)$ step-level overhead. Given token saliency scores, combining $T_i$ with its source-step utility adds $O(N_t)$ work. In our implementation, $T_i$ is collected during the existing observation prefill and requires no additional model pass.

The selector globally ranks $N_t$ token scores, with a worst-case cost of $O(N_t\log N_t)$. Therefore, StepKV adds $O(N_t\log N_t+t)$ time per pruning boundary, dominated by token ranking. Since every completed step contributes at least one trajectory token, $t\le N_t$, and the worst-case overhead simplifies to $O(N_t\log N_t)$.

\subsection{StepKV Pseudocode}
Algorithm~\ref{alg:stepkv} summarizes the overall procedure of StepKV.
After each reasoning step is completed, StepKV first computes the step utility,
then updates previous-step utilities based on later reuse signals. Finally,
it combines step-level utility with token-level saliency to rank candidate
tokens and retain the top tokens under the target cache budget.
\begin{algorithm}[t]
\DontPrintSemicolon
\caption{StepKV}
\label{alg:stepkv}
\KwIn{Prunable indices $\mathcal{I}^{(t)}$, finalized steps $\mathcal{S}^{(t)}$, keep ratio $\rho$}
\KwOut{Selected set $\mathcal{K}^{(t)}$ with $|\mathcal{K}^{(t)}|=B_t$}

\PhaseLine{Phase 1 (new-step scoring)}
Compute $\mathrm{valid}_t$, $\mathrm{gain}_t$, and $\mathrm{red}_t$ from $\mathrm{Th}_t$, $\mathrm{Act}_t$, and $\mathrm{Obs}_t$; compute $r_t$ via Equation~\eqref{eq:reward}\;
Set $c_{t}\leftarrow 0$ and compute $S_{t}$ via Equation~\eqref{eq:step-score}\;

\PhaseLine{Phase 2 (previous-step updating)}
\ForEach{previous step $1\le k<t$}{
    Update $c_k$ via Equation~\eqref{eq:reuse} and recompute $S_k$ via Equation~\eqref{eq:step-score}\;
}

\PhaseLine{Phase 3 (score and select)}
$B_t\leftarrow\max(1,\lfloor \rho|\mathcal{I}^{(t)}|\rfloor)$\;
For each token~$i$ produced by step~$k$, compute $P_i$ via Equation~\eqref{eq:token-priority}\;
Select the top $B_t$ tokens by $P_i$ as $\mathcal{K}^{(t)}$\;

\Return{$\mathcal{K}^{(t)}$}
\end{algorithm}

\section{Implementation Details}
\label{implementation}
\subsection{\textbf{Experiments Details.}}
We use Qwen2.5-7B-Instruct~\cite{qwen2.5} and Llama3.1-8B-Instruct~\cite{grattafiori2024llama} as backbone models. Full KV retains the complete cache without compression and serves as the 100\% cache reference, while ReAct denotes the original reasoning framework without KV cache budget constraints. All compressed methods use identical prompts, tools, decoding settings, maximum reasoning steps, pruning schedules, cache budgets, and random seeds to ensure a fair comparison. Field similarity is computed based on the overlap of up to 64 extracted evidence cues. Each experiment is repeated with three random seeds (233, 42, and 3407), and all reported results are presented as the mean $\pm$ standard deviation. Unless otherwise specified, StepKV uses the default hyperparameters listed in Table~\ref{tab:stepkv-hparams}.

\subsection{Prompt Template}
\begin{tcolorbox}[
    enhanced,
    breakable,
    colback=pink!3,             
    colframe=pink!40!black,     
    boxrule=0.6pt,
    arc=1.2mm,
    left=10pt,
    right=10pt,
    top=10pt,
    bottom=10pt,
    title={\textbf{ReAct Prompt Template}},
    fonttitle=\bfseries\large,
    attach boxed title to top left={yshift=-2mm}
]

{\color{pink!60!black}\textbf{Instruction}}

Solve a question answering task with interleaving \Thought{}, \Action{}, and \Observation{} steps. \Thought{} can reason about the current situation, and \Action{} can be three types:

(1) Search[entity], which searches Wikipedia and returns the first paragraph if it exists.

(2) Lookup[keyword], which returns the next sentence containing keyword.

(3) Finish[answer], which returns the final answer and terminates the task.

\vspace{2mm}

{\color{pink!60!black}\textbf{Few-shot Example}}

\texttt{Question: Musician and satirist Allie Goertz wrote a song about Milhouse, who was he named after?}

\texttt{Thought 1: I should search Milhouse.}

\texttt{Action 1: Search[Milhouse]}

\texttt{Observation 1: Milhouse is a character in The Simpsons.}

\texttt{Thought 2: I should look up naming origin.}

\texttt{Action 2: Lookup[named after]}

\texttt{Observation 2: Milhouse was named after Richard Nixon.}

\texttt{Action 3: Finish[Richard Nixon]}

\end{tcolorbox}

\begin{table}[H]
\centering
\caption{Implementation hyperparameters and default settings for StepKV.}
\label{tab:stepkv-hparams}
\footnotesize
\setlength{\tabcolsep}{2pt}
\renewcommand{\arraystretch}{1.08}
\begin{tabular}{@{}
>{\raggedright\arraybackslash}p{0.24\columnwidth}
>{\raggedright\arraybackslash}p{0.18\columnwidth}
>{\raggedright\arraybackslash}p{0.46\columnwidth}
@{}}
\toprule
Parameter & Default & Role \\
\midrule
cue cap & 64 & max evidence cues per field \\
$\lambda$ & $0.3$ & redundancy penalty in $r_k$\\
$w_r, w_c$ & $0.85, 0.15$ & completion / reuse weights \\
$r_{\min}, r_{\max}$ & $-1, 2$ & completion-score clipping range \\
$S_{\max}$ & $8$ & utility clipping bound \\
$\eta$ & $1.5$ & reuse update scale \\
$\alpha, \beta$ & $0.2, 0.8$ & token / step weights \\
$\rho$ & $0.2$--$0.5$ & global KV keep ratio\\
saliency layers & last $3$ layers & layers used for token saliency\\
\bottomrule
\end{tabular}
\end{table}

\section{Additional experiments analysis}

\begin{figure*}[t]
    \centering
    \begin{subfigure}[c]{0.34\textwidth}
        \centering
        \includegraphics[width=0.88\linewidth]{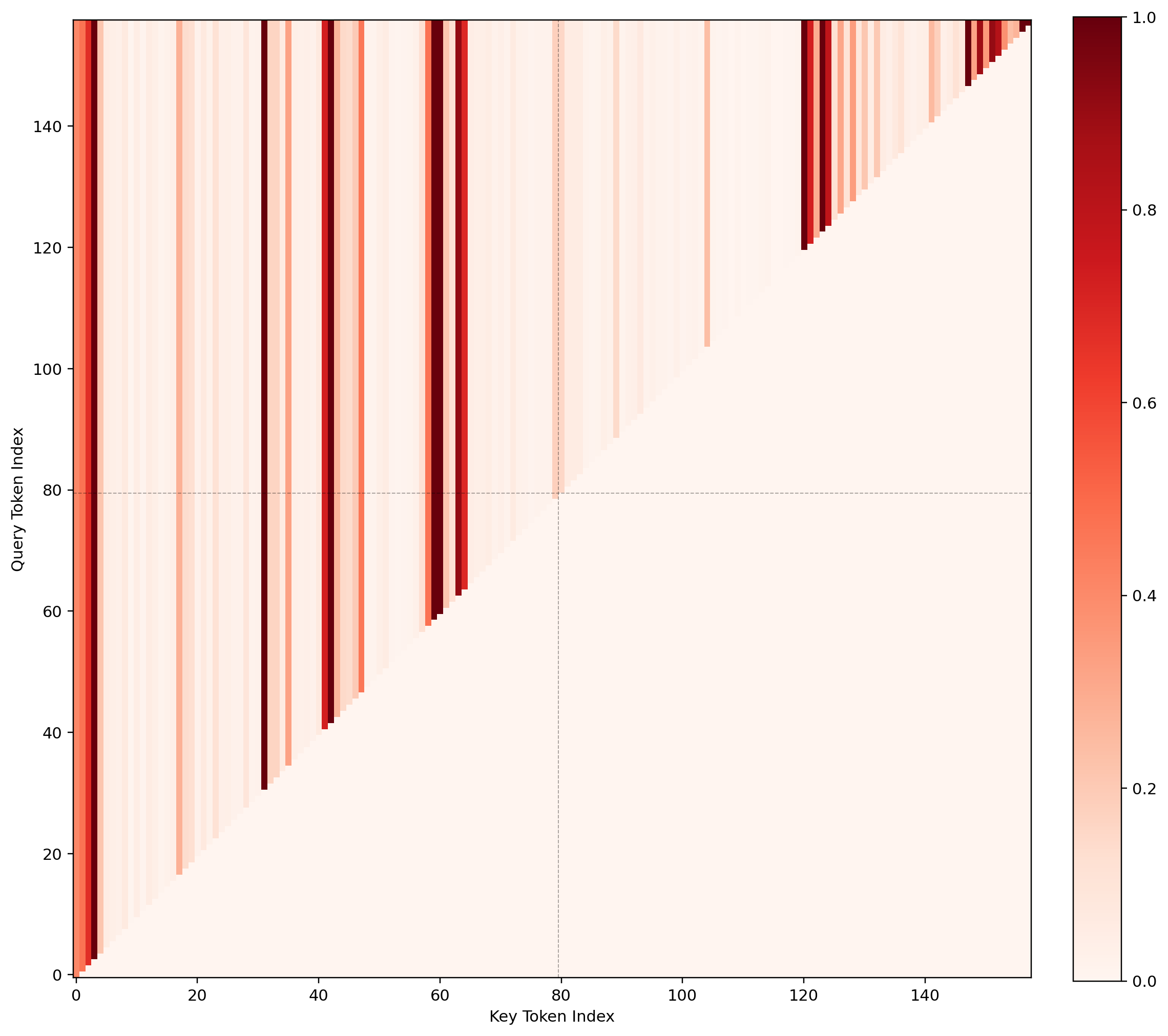}
        \caption{Aggregate HotpotQA retention pattern.}
        \label{fig:hotpot_analysis}
    \end{subfigure}\hfill
    \begin{subfigure}[c]{0.62\textwidth}
        \centering
        \includegraphics[width=\linewidth]{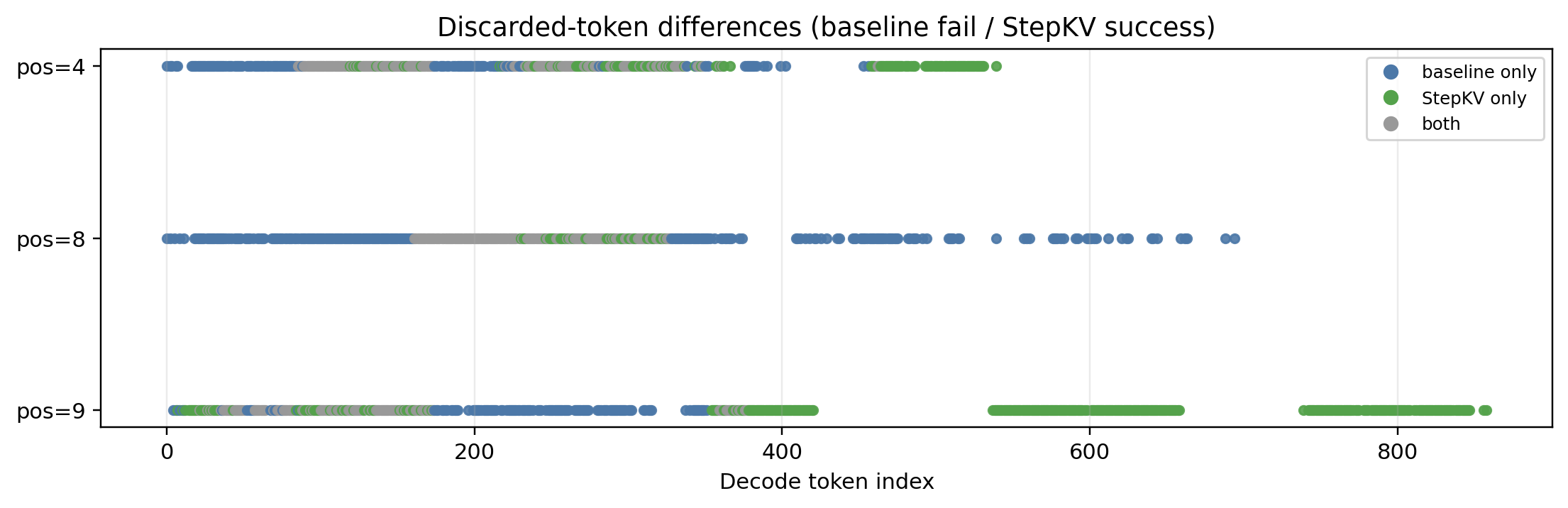}
        \caption{Baseline failure and StepKV success.}
        \label{fig:baseline_failure_case}
    \end{subfigure}
    \caption{Step-aware retention at two levels of analysis. Panel (a) shows the aggregate token-retention structure under cache compression, while panel (b) compares a baseline failure with a successful StepKV trajectory.}
    \label{fig:retention-mechanism}
    \Description{Aggregate and trajectory-level analyses of how StepKV changes retained context.}
\end{figure*}

\textbf{Step-aware retention changes the retained context at both aggregate and trajectory levels.} Figure~\ref{fig:retention-mechanism}(\subref{fig:hotpot_analysis}) visualizes the HotpotQA token-retention structure associated with the aggregate gains in Table~\ref{tab:main_results}. Figure~\ref{fig:retention-mechanism}(\subref{fig:baseline_failure_case}) then shows the same mechanism in an individual trajectory: token-level pruning removes intermediate evidence required by later reasoning, whereas StepKV preserves the relevant spans and completes the task successfully.

\begin{figure}[t]
    \centering
    \includegraphics[width=\linewidth]{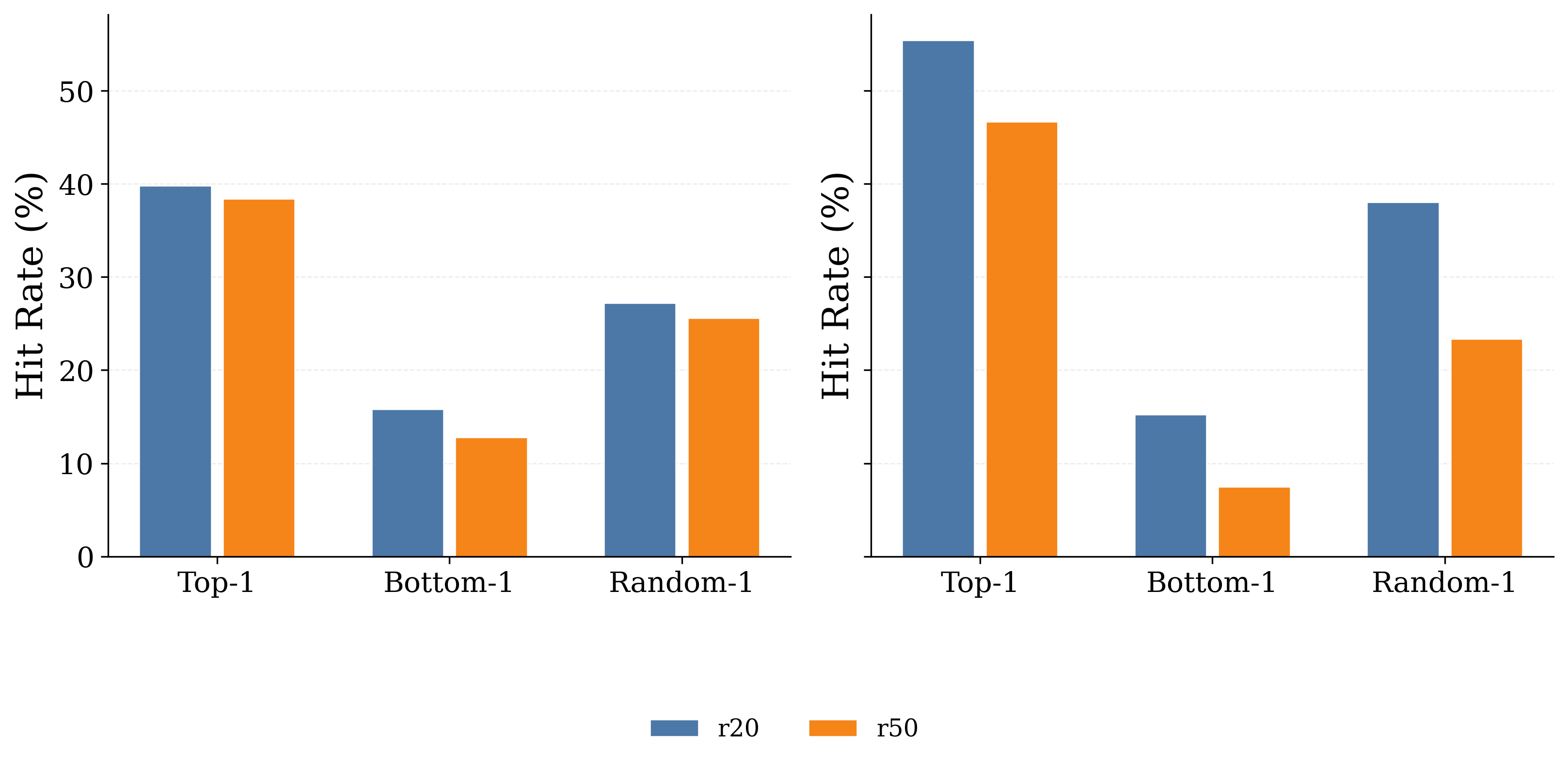}
    \caption{Offline analysis of step contribution using answer-relevant keyword overlap.}
    \label{fig:offline_step_contribution}
\end{figure}

\textbf{Estimated step importance aligns with answer-relevant evidence.} Figure~\ref{fig:offline_step_contribution} compares high-scoring, low-scoring, and random steps using answer-relevant keyword overlap. High-scoring steps contain more answer-relevant content, supporting the design choice of assigning different retention pressure to different reasoning steps.

\begin{figure}[h]
    \centering
    \includegraphics[width=\linewidth]{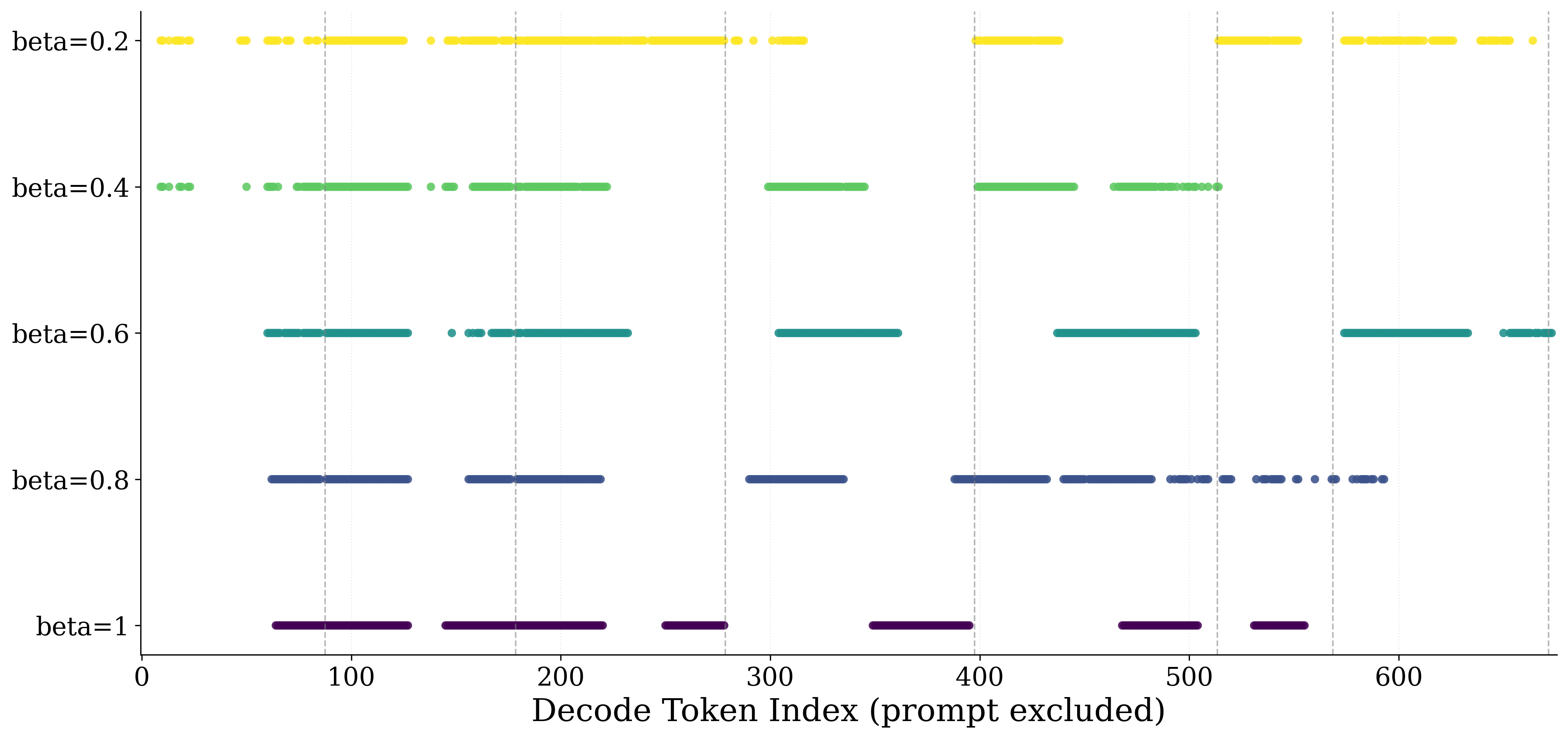}
    \caption{Token-retention patterns under different step-score mixing weights.}
    \label{fig:beta_sensitivity}
\end{figure}

\textbf{The step-score weight changes the retained trajectory structure, not only the final metric.} Figure~\ref{fig:beta_sensitivity} visualizes token-retention patterns under different mixing weights, showing that the hyperparameter alters which parts of the trajectory remain available during decoding. This explains why the performance sensitivity in Figure~\ref{fig:hyper_parameter} is tied to a real change in cache contents.

\section{Limitations}

Our current implementation estimates step utility using lightweight trajectory-derived signals, which provides an efficient and training-free solution but may introduce estimation uncertainty in challenging scenarios. For example, the quality of these signals can be affected when action-observation grounding is incomplete, tool responses are noisy, or important information is implicitly expressed and difficult to extract. Moreover, our evaluation is conducted on a limited range of model scales and agent benchmarks. Future work will extend the evaluation of StepKV to larger-scale language models and more diverse agent tasks to further investigate its generalization and scalability. Another promising direction is to develop more accurate and effective step utility estimation methods by exploiting richer internal signals, such as step-level representations, hidden states, or other model-intrinsic information, which may provide a more precise measurement of reasoning-step importance beyond external trajectory signals.



\end{document}